\documentclass[sigconf]{aamas} 

\usepackage{balance} 

\setcopyright{ifaamas}
\acmConference[AAMAS '26]{Proc.\@ of the 25th International Conference
on Autonomous Agents and Multiagent Systems (AAMAS 2026)}{May 25 -- 29, 2026}
{Paphos, Cyprus}{C.~Amato, L.~Dennis, V.~Mascardi, J.~Thangarajah (eds.)}
\copyrightyear{2026}
\acmYear{2026}

\newcommand{\BibTeX}{\rm B\kern-.05em{\sc i\kern-.025em b}\kern-.08em\TeX}
\newcommand{\Safe}{\textsc{safe}}
\newcommand{\Unsafe}{\textsc{unsafe}}

\acmSubmissionID{TBD}

\title[]{Humans Are More Diverse: Frontier LLMs Show Extreme Policies in Idealised AI Development Races}

\newcommand{\HCMUSaffiliation}{%
  \affiliation{%
    \institution{Ho Chi Minh City University of Science}
    \institution{Vietnam National University Ho Chi Minh City (VNU-HCM)}
    \city{Ho Chi Minh City}
    \country{Vietnam}
  }%
}

\newcommand{\HCMUTaffiliation}{%
  \affiliation{%
    \institution{Ho Chi Minh City University of Technology (HCMUT)}
    \institution{Vietnam National University Ho Chi Minh City (VNU-HCM)}
    \city{Ho Chi Minh City}
    \country{Vietnam}
  }%
}

\newcommand{\BrusselsAffiliation}{%
  \affiliation{%
    \institution{Vrije Universiteit Brussel}
    \institution{Universit\'e Libre de Bruxelles}
    \city{Brussels}
    \country{Belgium}
  }%
}

\newcommand{\TeessideAffiliation}{%
  \affiliation{%
    \institution{School of Computing, Engineering and Digital Technologies}
    \institution{Center for Digital Innovation, Teesside University}
    \city{Middlesbrough}
    \country{United Kingdom}
  }%
}

\author{Phu-Hoa Pham}
\authornotemark[1]
\HCMUSaffiliation
\email{23122030@student.hcmus.edu.vn}

\author{Duy-Minh Dao-Sy}
\authornote{These authors contributed equally to this work.}
\HCMUSaffiliation
\email{23122041@student.hcmus.edu.vn}

\author{Trung-Kiet Huynh}
\authornotemark[1]
\HCMUSaffiliation
\email{23122039@student.hcmus.edu.vn}

\author{Phu-Quy Nguyen-Lam}
\authornotemark[1]
\HCMUSaffiliation
\email{23122048@student.hcmus.edu.vn}

\author{Chi-Nguyen Tran}
\authornotemark[1]
\HCMUSaffiliation
\email{23122044@student.hcmus.edu.vn}
\author{Minh-Trung Le}
\authornotemark[1]
\HCMUTaffiliation
\email{trung.leminh@hcmut.edu.vn}

\author{Phong-Hao Le}
\authornotemark[1]
\HCMUTaffiliation
\email{hao.lephong@hcmut.edu.vn}

\author{Dinh-Nam Nguyen}
\authornotemark[1]
\HCMUTaffiliation
\email{nam.nguyenkhmt@hcmut.edu.vn}

\author{Thien-Ky Nguyen-Dong}
\authornotemark[1]
\HCMUTaffiliation
\email{ky.nguyendongthien@hcmut.edu.vn}

\author{Elias Fern\'andez Domingos}
\BrusselsAffiliation
\email{elias.fernandez.domingos@vub.be}

\author{Le Hong Trang}
\authornote{Corresponding authors.}
\HCMUTaffiliation
\email{lhtrang@hcmut.edu.vn}

\author{The Anh Han}
\authornotemark[2]
\TeessideAffiliation
\email{T.Han@tees.ac.uk}

\begin{abstract}
An AI development race creates a multi-agent safety dilemma. Each company can
develop slowly and safely, or move faster while taking a risk that may remove
its final reward. We use this repeated game to study strategic safety behaviour
among large language model (LLM) agents in races with two to five players.
However, a valid action does not show that an agent understands the game. We
therefore place an audit gate before behavioural interpretation.
We first verify the game engine, then test rule recall, state tracking, payoff
calculation, and stability under different but equivalent task descriptions.
We then compare LLM action sequences with an evolutionary game-theory
benchmark and published human data, and explore differences across models,
risk conditions, personas, and two- to five-player races. The audit shows
that strong rule recall can
coexist with weak state tracking and expected-payoff calculation. Providing
verified arithmetic and changing the response representation can also change
later actions, even when the game rules stay fixed. Across seven tested model
endpoints, aggregate rates hide large differences in action sequences,
responses to opponents, and responses to race position. Patterns across the
tested three- to five-player races are also model-specific rather than a single
effect of adding competitors. These results show why multi-agent AI-race
simulations need validity checks and trajectory-level analysis before their
outputs are described as strategic, human-like, or safety-aware. Our findings
are exploratory and apply only to the tested models, prompts, and decoding
settings.
\end{abstract}


\renewcommand\footnotetextcopyrightpermission[1]{}
\begin{document}


\pagestyle{fancy}
\fancyhead{}


\maketitle 


\section{Introduction}
\label{sec:introduction}

Competition over advanced artificial intelligence can create a safety
dilemma. Each company may benefit from moving faster, but safety work can slow
it down when its rivals do not follow the same standard
\citep{ai_race_for_strategic_advantage,racing_to_the_precipice,role_of_Cooperation}.
An \emph{AI development race} is a simple game that represents this tension.
In each round, every player chooses between slower Safe development and faster
Unsafe development \citep{to_regulate_or_not,artificial_intelligence_development_races}.
The game is idealised: it does not model a real AI company in full. Its value
is control. It lets us test how choices change with competition, relative
progress, and the earlier actions of other players.

Human experiments show why the order of actions matters. In the two-player
study of \citet{falling_behind_unsafe}, the preregistered comparison between
the two higher private-risk conditions found no clear difference. Exploratory
analyses instead linked later Unsafe choices to three parts of the race
history: the opponent's previous action, whether the player was ahead or
behind, and the player's first action. We call this \emph{path dependence}:
an early choice or event changes the later path of the interaction. A useful
human--LLM comparison must therefore examine full action sequences, which we
call \emph{trajectories}, rather than only the overall Unsafe rate.

Large language models (LLMs) can act as language-based players in this game.
They are already studied as simulated participants, decision-making agents,
and game-playing systems
\citep{using_llm_to_simulate,out_of_one,playing_repeated_games_with_llms}.
Yet their use creates a measurement problem. A model may return a correctly
formatted action without correctly tracking the state or calculating the
payoff. In a repeated game, one such action changes the next state and can
affect every later round. A plausible final trajectory is therefore not, by
itself, evidence that the model understood the game.

Prompt-based evaluation is itself sensitive to presentation choices that do
not necessarily change task meaning. Meaning-preserving few-shot formatting
changes can produce large, model-dependent performance spreads
\citep{sclarPromptFormatting2024}; reordering answer options can change
multiple-choice predictions \citep{pezeshkpourOptionOrder2024}; and order
and response-token identity can create distinct selection biases
\citep{weiSelectionBias2024}. Controlled studies also report answer changes
under output-format requests and atomic whitespace variants
\citep{salinasButterfly2024}. Persona prompts should therefore be treated as
experimental interventions rather than generic performance improvements:
across a large factual-question evaluation, their average benefit was absent
or slightly negative and the best persona was difficult to select reliably
\citep{zhengPersona2024}. These findings motivate our separate audits of
wording, answer order, arithmetic disclosure, narrative framing, and opaque
response mapping.

Game-specific evaluations reveal the same validity threat in strategic
choice. In Stag Hunt and Prisoner's Dilemma presentations, reversing the
order of action labels and changing their payoff assignments altered choice
frequencies, with model-specific positional and payoff biases
\citep{herrStrategicBias2024}. Across procedurally generated Prisoner's
Dilemma vignettes, decisions varied substantially with topic, actor, and
world framing despite a common underlying game structure
\citep{robinsonBurdenFraming2025}. More generally, performance across 11
counterfactual tasks declined consistently when familiar default mappings
were replaced by explicitly specified alternatives
\citep{wuCounterfactual2024}. These results support our decision to test
action-code mappings and narrative skins independently, and to avoid treating
one familiar presentation as a neutral measurement instrument.

Reproducible game-based frameworks provide a complementary way to study these
effects. FAIRGAME standardises repeated-game experiments across models,
languages, and persona conditions, and supports comparison with game-theoretic
predictions \citep{buscemiFAIRGAME2025}. Later work extends this approach with
a payoff-scaled Prisoner's Dilemma, a three-player Public Goods Game, and
supervised recognition of canonical strategies from action trajectories
\citep{huynhUnderstanding2025}. A related payoff-scaling study finds that
cooperation and inferred strategy distributions vary across models, languages,
and incentive magnitudes, and can move in the opposite direction from its
evolutionary benchmark \citep{huynhPayoffScaling2026}. These studies motivate
our use of full trajectories and theoretical benchmarks, while our audit asks
the separate question of whether an agent can correctly track the race that
produces those trajectories.

Behavioural evaluations that focus on action frequencies or strategic
outcomes can obscure this distinction unless task validity and
representation robustness are established separately. An LLM may produce
an aggregate \Unsafe{} rate similar to that of human participants while
relying on incorrect state computations, prompt-specific heuristics, or
unstable action encodings. Conversely, two checkpoints with similar
aggregate action rates may exhibit substantially different responses to
opponent behaviour, relative position, or early-round events. Aggregate
behavioural similarity is therefore insufficient evidence that an LLM
understands the task, reproduces human decision dynamics, or exhibits a
general safety preference \citep{synthetic_replacement_human}.

We address this problem with an \emph{audit-first evaluation}: we check that
the game and the agent's use of its rules are valid before we interpret the
agent's behaviour. We reproduce the two-player game studied by
\citet{falling_behind_unsafe} and add a compatible multiplayer version based
on \citet{to_regulate_or_not}. All agents in a round see the same state before
anyone acts. Their choices remain hidden until every agent has responded, so
the choices are simultaneous. The game engine, not the LLM, then calculates
progress, payoffs, the end of the race, prizes, ties, and setbacks. We save
every prompt, response, parsed action, retry, state change, configuration,
random seed, and final outcome.

The audit has four levels. First, \emph{mechanical validity} means that the
game engine applies the stated rules correctly. Second, \emph{task validity}
means that an agent can recall the rules, rebuild the current state, apply a
state change, and calculate the final result. Third, \emph{representation
robustness} means that choices remain stable when wording or response labels
change but the game itself does not. Fourth, we compare valid trajectories
with published human patterns. The human data are a reference for observable
behaviour. They are not a target that an LLM must copy, and LLM outputs are
not treated as replacements for human participants.

That boundary is supported by mixed evidence from LLM-based behavioural
simulation. Aher et al. reproduced qualitative findings in three of four
experiments but found a hyper-accuracy distortion in the fourth
\citep{using_llm_to_simulate}. Repeated-game work further shows that a model
may predict an opponent's pattern without immediately acting in accordance
with that prediction, while eliciting the prediction before action can
improve coordination and score \citep{playing_repeated_games_with_llms}.
Accordingly, neither a plausible trajectory nor a verbal strategy report is
treated here as sufficient evidence of task comprehension.

Distributional agreement is also stricter than matching a treatment mean.
Dynamic market simulations have recovered broad differences between positive-
and negative-feedback conditions while exhibiting less strategy
heterogeneity than human participants \citep{delRioChanonaMarkets2025}.
Accordingly, our human comparison reports trajectory archetypes and
within-population diversity alongside aggregate \Unsafe{} rates.

This emphasis on distributions rather than plausible prose is shared by
broader evaluation work. CogBench phenotypes models with behavioural metrics
from seven cognitive experiments \citep{codaFornoCogBench2024}, while
SocioBench reports substantial individual- and subgroup-level gaps across
large international survey data \citep{wangSocioBench2025}. Persona-based
simulation can reproduce linguistic patterns without recovering the full
interaction of human perceptions \citep{wangHumanSubjectivity2024}, and
specialised survey simulators still struggle on unseen questions
\citep{suhSurveyDistributions2025}. Position papers consequently frame LLM
social simulation as most defensible for pilot and exploratory use unless it
is validated against the relevant population \citep{anthisSocialSimulation2025}.
Strategic-game evidence is similarly heterogeneous: repeated Prisoner's
Dilemma policies are not fully consistent across parameter and framing
changes \citep{palCooperation2026}; human-like heuristics can be applied more
rigidly than by human participants \citep{zhengBoundedRationality2025};
multi-round agents may cooperate instead of converging to the predicted Nash
equilibrium \citep{yaoCompetition2026}; and real multi-turn action sequences
remain difficult to reproduce from prompts alone \citep{luMultiTurn2026}.

We organise the paper around three questions about strategic behaviour in an
AI race:
\begin{itemize}
    \item \textbf{RQ1 - Strategic behaviour in two-player races:}
    How do agents respond to risk, opponent history, relative progress, and
    early choices, compared with game-theory and human benchmarks?

    \item \textbf{RQ2 - Multi-agent race dynamics:}
    Across races with three to five agents, how is Unsafe play associated with
    rank, model, persona, and tested group size?

    \item \textbf{RQ3 - Validity and strategic robustness:}
    Can agents track the game well enough for strategic interpretation, and
    are choices stable across equivalent task presentations?
\end{itemize}

The main insight is that the tested LLMs do not express one common strategic
policy. In two-player races, models with similar aggregate Unsafe rates can
respond differently to risk, opponent history, and relative progress. In
three- to five-player races, the association between rank and Unsafe play can
change with both model and persona, while aggregate behaviour does not move
monotonically with group size. These are exploratory multi-agent patterns, not
an identified causal effect of adding one more player. The validity audit also
sets an important boundary: rule recall can coexist with incorrect state
tracking, and an equivalent representation of the same game can change later
actions.

The novelty lies in connecting four views of one AI race. We provide an
auditable mechanism for two- and multi-player races; compare LLM play with
game-theory and human benchmarks; extend the analysis to three to five agents;
and use a validity gate before strategic interpretation. Together, these views
show model-specific response rules and multi-agent patterns that aggregate
Unsafe rates hide, while keeping diagnostic and confirmatory evidence separate.

Each reported behavioural conclusion is scoped to its tested checkpoint,
prompt version, decoding contract, experimental configuration, and run
period. The labels \Safe{} and \Unsafe{} denote actions within the experimental game and should not be interpreted as direct measurements of general LLM safety, subjective risk preference, or suitability for autonomous deployment.

\section{Preliminaries}
\label{sec:preliminaries}

\begin{figure*}[t]
  \centering
  \includegraphics[width=\textwidth]{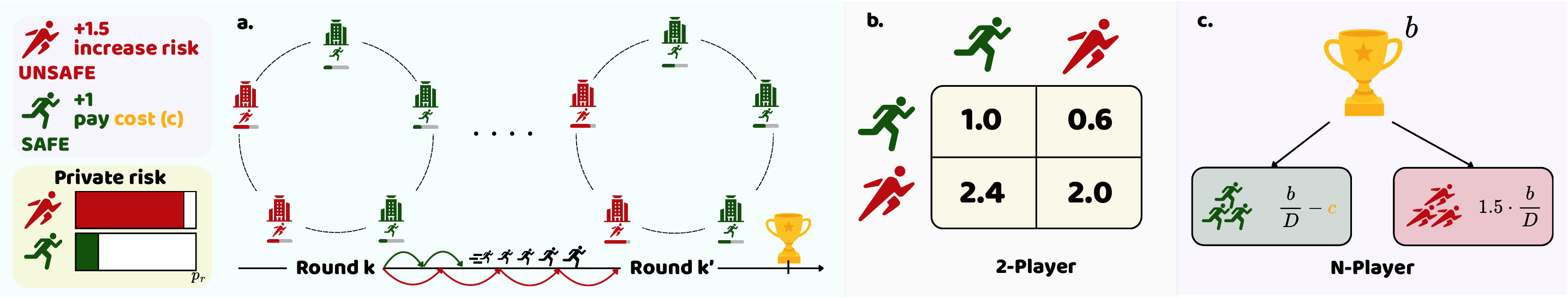}
  \caption{Overview of the AI-race mechanism shared by the two-player and $N$-player games. (a) Each round, every company advances along the track by choosing Safe (green, $+1$ progress, pays cost $c$) or Unsafe (red, $+1.5$ progress, increases private setback risk); the race ends after a hidden number of rounds, at which point terminal prize $B$ is awarded to the progress leader or divided among tied leaders. (b) The two-player stage-payoff matrix: own Safe/opponent Safe pays 1.0, own Safe/opponent Unsafe pays 0.6, own Unsafe/opponent Safe pays 2.4, and own Unsafe/opponent Unsafe pays 2.0. (c) The $N$-player stage-payoff rule: if $k$ of $N$ companies choose Safe, $D=k+s(N-k)$; each Safe-choosing company earns $b/D-c$ and each Unsafe-choosing company earns $s b/D$, where $b$ is the per-round benefit and is distinct from terminal prize $B$.}
  \label{fig:mechanism}
  \Description{Three-panel diagram of the AI-race mechanism. Panel a shows company icons moving around a circular track, with green Safe icons advancing one step and red Unsafe icons advancing one and a half steps while increasing a private risk bar; the track leads to terminal prize B after an unmarked number of rounds. Panel b shows the two-player payoff matrix as a two-by-two grid labelled Safe and Unsafe on each axis, with values 1.0, 0.6, 2.4, and 2.0. Panel c shows the $N$-player stage-payoff rule: if $k$ of $N$ companies choose Safe, $D=k+s(N-k)$; Safe receives $b/D-c$ and Unsafe receives $s b/D$. Lowercase $b$ denotes per-round benefit, not terminal prize $B$.}
\end{figure*}

We first specify the two-player game reproduced from \citet{falling_behind_unsafe} (\S\ref{sec:prelim-2p}), then its generalisation to $N>2$ companies (\S\ref{sec:prelim-np}). Every quantity that is not restated in \S\ref{sec:prelim-np}, namely the horizon, prize split, and private-risk formula, carries over unchanged from the two-player game; only the stage payoff changes shape (Figure~\ref{fig:mechanism}).


\subsection{Two-player game}
\label{sec:prelim-2p}

Two companies $i \in \{1,2\}$ play a repeated race. In every round $t$, both companies simultaneously choose an action $a_i^t \in \{S, U\}$ (Safe, Unsafe) and commit to it before either learns the other's choice for that round; only once both actions are committed does the round resolve (Figure~\ref{fig:mechanism}a). Resolution has two parts. First, Safe advances the company's cumulative progress by $\sigma(S)=1.0$ and Unsafe advances it by $\sigma(U)=1.5$, so Unsafe always closes the race faster; progress accumulates as $P_i^t = P_i^{t-1}+\sigma(a_i^t)$. Second, the round pays a stage payoff $\pi_i^t=\pi(a_i^t,a_{-i}^t)$, read off the fixed matrix (own action by row, opponent action by column, Figure~\ref{fig:mechanism}b):
\begin{equation}
    \pi = \begin{pmatrix} 1.0 & 0.6 \\ 2.4 & 2.0 \end{pmatrix}.
    \label{eq:2p-matrix}
\end{equation}
Unsafe strictly dominates Safe at the stage-game level (row 2 exceeds row 1 in both columns): the only cost of Unsafe development enters later, through accumulated private risk rather than through the round payoff itself.

The race lasts a minimum of $5$ rounds; after every completed round from round $5$ onward it terminates with probability $0.2$, giving an expected horizon $\mathbb{E}[T]=9$. Neither company is ever told the horizon in advance, so a company can never condition its action on knowing the current round is the last one (Figure~\ref{fig:mechanism}a). Termination then resolves in two further steps. First, the company with the higher cumulative progress $P_i^T$ is declared the winner of a prize $B=100$; if progress is exactly tied, the two companies split it $50/50$. Second, and only for that winner or tied winners, an accumulated private setback risk is drawn:
\begin{equation}
    q_i(T) = p_r^{\max} \cdot \frac{n_i^U(T)}{T}, q_i(T) \in [0,1], p_r^{\max}\in\{0.10,0.60,0.90\},
    \label{eq:risk}
\end{equation}
where $n_i^U(T)$ counts company $i$'s Unsafe choices over the whole race and $p_r^{\max}$ is a treatment-level constant, assigned once per race, that we vary across three risk levels. If the draw against $q_i(T)$ fails, that company's entire race payoff, every stage payoff it accumulated plus the prize, is zeroed; if it succeeds, the company keeps both. A company that ends behind is never subject to this draw at all: it keeps its accumulated stage payoffs regardless of how much Unsafe it played, but receives no share of the prize (Figure~\ref{fig:mechanism}a). This asymmetry, in which risk is built by both companies but realised only against whoever is ahead, is what makes Unsafe development a genuine gamble rather than a free speed advantage.

\subsection{$N$-player game}
\label{sec:prelim-np}

The $N$-player game keeps every element of \S\ref{sec:prelim-2p} unchanged (simultaneous sealed actions, progress increments $\sigma(S)=1.0$/$\sigma(U)=1.5$, the minimum-5-round hidden horizon, terminal prize $B=100$, and the risk formula, Equation~\ref{eq:risk}), but is played by $N\geq2$ companies indexed by $i\in\{1,\dots,N\}$ instead of exactly two, and the prize at round $T$ splits evenly among however many companies share the highest progress rather than only two. The engine supports \(N\in\{2,3,4,5\}\). The two-player condition
reproduces the original human-study mechanism, while the current
multiplayer experiments cover matched conditions with
\(N\in\{3,4,5\}\) (Figure~\ref{fig:mechanism}c).

With more than two companies there is no single opponent to index a matrix against, so Equation~\ref{eq:2p-matrix} generalises to the $N$-player stage-payoff rule of \citet{to_regulate_or_not} (their Appendix~B), evaluated at $k^t \in \{0,1,\dots,N\}$, the number of companies (out of $N$) choosing Safe in round $t$. Therefore, every Safe-choosing company earns
\begin{equation}
    \pi_{\text{Safe}}(k^t) = \frac{b}{k^t + s\,(N-k^t)} - c,
    \label{eq:np-safe}
\end{equation}
and every Unsafe-choosing company earns
\begin{equation}
    \pi_{\text{Unsafe}}(k^t) = s \cdot \frac{b}{k^t + s\,(N-k^t)},
    \label{eq:np-unsafe}
\end{equation}
with cost $c=1$, per-round benefit $b=4$, and speed $s=1.5$ fixed across all group sizes (the per-round benefit $b$ is distinct from terminal prize $B$, which is paid once, at the end of the race, not every round). Both payoffs depend on the \emph{group's} joint action counts, $k^t$ and $N-k^t$, not on any single rival's action, which is the structural difference from the two-player game: Thus, stage payoffs depend on the number of rivals choosing each action,
not on their identities.

\subsection{Game-theoretic benchmark}
\label{sec:theory-benchmark}

The game contains two opposing incentives. In a single round, Unsafe strictly
dominates Safe: it gives more progress and a higher stage payoff against either
action. Across the full race, however, Unsafe choices increase the setback risk
for a winner or tied winner. The best action can therefore depend on the risk
condition, earlier actions, and the expected response of other players. This is
the strategic tension that the LLM agents must navigate.

We use four simple strategies as a theory benchmark: Always Safe (AS), Always
Unsafe (AU), Conditional Safe (CS), which starts Safe and then copies the
opponent, and Conditional Unsafe (CAS), which starts Unsafe and then copies the
opponent. In the reconstructed finite-population evolutionary model of
\citet{falling_behind_unsafe}, the most common strategy changes from AU at
$p_r^{\max}=0.10$, to CAS at $0.60$, and to CS at $0.90$. Thus, the theory does
not predict one fixed amount of Unsafe play. It predicts a risk-dependent shift
between unconditional speed, an aggressive opening, and conditional restraint.
We use this as a qualitative benchmark, not as a fitted model of LLM output:
prompted self-play trajectories are not samples from an evolutionary population.

\section{Experimental design and evidence policy}
\label{sec:design}
\begin{figure*}[t]
  \centering
  \includegraphics[width=0.7\textwidth]{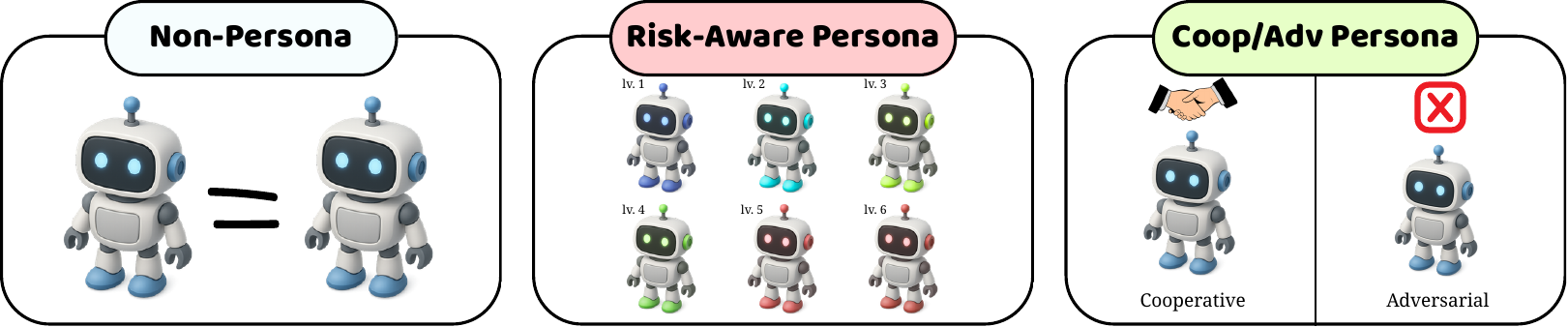}
  \caption{Overview of the three persona-related framing families used to
  describe each company's executive. The canonical baseline contains no
  persona sentence; a separate neutral placebo uses a length-matched neutral
  sentence. The risk-aware family contains six levels adapted from the
  Eckel--Grossman gamble scale. The cooperative/adversarial family frames
  each seat as believing rival firms do better through mutual restraint or
  only at each other's expense. These are prompt conditions, not measured
  psychological traits.}
  \label{fig:persona}
  \Description{Three-panel illustration using simple robot mascots to
  represent company executives. Panel 1 shows the non-persona condition.
  Panel 2 shows six risk-aware levels. Panel 3 contrasts cooperative and
  adversarial framing.}
\end{figure*}
\subsection{Agent protocol and units of analysis}

At each decision, an agent receives a versioned prompt containing the current round, assigned maximum private risk, the publicly disclosed state, and the preceding revealed action profile when $t>1$. Same-round decisions are sealed: every response is obtained from the same pre-action snapshot before the engine resolves actions, progress, payoffs, and risk. The environment, rather than generated text or arithmetic, determines all transitions. Raw response, parsed action, retry count, parser status, pre-turn state, terminal record, configuration, and seed allocation are retained for every decision.

The main behavioural outcome is whether an agent chooses Unsafe. The race,
not the individual decision, is the main independent experimental unit.
Decisions from the same race depend on earlier states and therefore cannot be
treated as separate random samples. We report the number of races,
trajectories, and decisions. Where possible, uncertainty is grouped by race
or matched repetition. A \emph{parse failure} occurs when a response does not
follow the required action format. Because the resulting fallback action
changes later states, one parse failure marks the whole race as contaminated.
\subsection{Evidence strata and admission}

We separate mechanical validity, task validity, diagnostic pilots, and
confirmatory evidence. A \emph{diagnostic pilot} is a small exploratory test
used to find possible effects or failures. It does not estimate a stable
population effect. A \emph{confirmatory} study tests a question under a plan
fixed before its results are inspected. A result enters that stronger evidence
level only when the protocol, model identifier, prompt and configuration
hashes, completed-race count, and exclusion rules were all fixed and recorded.

The primary audit uses protocol ai-race-game-understanding-v2: 41 atomic probes spanning rule recall, one-stage payoffs, state reconstruction, state transition, terminal scoring, and expected payoff. Numeric probes vary direct wording, paraphrase, and calculator disclosure; categorical probes also reverse answer order. Strict output-format compliance and semantic correctness are scored separately. A paired behavioural diagnostic compares the canonical prompt with a deterministic four-row decision card that discloses the focal agent's immediate payoff, progress, and private-risk consequences for each action profile. The card does not reveal the opponent's move or the stopping horizon.

We also vary the persona-related framing assigned to each company's
executive. The canonical baseline contains no persona sentence, while a
separate neutral placebo uses a length-matched neutral description. The
risk-aware family contains six levels adapted from the Eckel--Grossman gamble
scale, and the cooperative/adversarial family frames the executive as favouring
either mutual restraint or zero-sum competition. These are prompt conditions,
not measured psychological traits (Figure~\ref{fig:persona}).

The context extension keeps the game rules fixed but presents the game through
eight different stories, which we call \emph{narrative skins}. It also replaces
Safe and Unsafe with the opaque response codes $P$ and $Q$. We study this in
three ways. A paired first-round comparison tests the same initial state. A
\emph{fixed-state replay} asks for a new action at a previously recorded state
without allowing that answer to change later states. It therefore measures the
direct effect of the prompt at that state. A \emph{live trajectory} lets each
answer change the next state, so direct prompt effects and later feedback can
build on each other. A separate comprehension test acts as an admission gate.
If that gate fails, a behavioural difference shows prompt-conditioned output,
not informed optimisation of the game.

\begin{table}[t]
  \caption{Evidence strata used in this paper. Pilot modules are never pooled with confirmatory inference.}
  \label{tab:evidence-strata}
  \centering
  \footnotesize
  \setlength{\tabcolsep}{3pt}
  \begin{tabular}{@{}p{0.29\columnwidth}rrp{0.23\columnwidth}@{}}
    \toprule
    Module & Races & Decisions & Use \\
    \midrule
    Qwen task audit & 685 & -- & validity audit \\
    Calculator diagnostic & 60 & 1,116 & paired diagnostic \\
    Context pilot ($T=0$) & 768 & 13,680 & robustness diagnostic \\
    Fixed-state replay & 1,536 & -- & direct prompt diagnostic \\
    Five-checkpoint baseline & 150 & 2,790 & descriptive only \\
    OpenAI $N=3$--$5$ baseline & 180 & 6,120 & exploratory only \\
    \bottomrule
  \end{tabular}
\end{table}

\section{Results}
\label{sec:results}

We report the validity gate first because every strategic claim depends on it.
We then answer RQ1 by comparing two-player LLM trajectories with the
evolutionary and human benchmarks, including responses to risk, opponent
history, and relative progress. We answer RQ2 with the three- to five-player
pilots, focusing on rank and persona. Representation diagnostics answer RQ3
and define where the behavioural interpretation must stop. The Appendix
contains full model tables, predictive diagnostics, and robustness checks.

\subsection{Validity gate: can agents follow the race?}
\label{sec:results-validity}

We first ask whether the audited model can carry out the rules it receives.
The admitted audit contains 685 fixed-seed outputs from
Qwen2.5-7B-Instruct, and all planned outputs were retained. As
Table~\ref{tab:task-audit} shows, the model recalls rules and looks up a
one-round payoff almost perfectly. It performs much worse when it must track
the changing state. It follows the requested one-line format in only 32.1\%
of outputs, although the frozen semantic parser can recover a value from
every response. Showing a verified calculator result also raises accuracy
from 52.1\% to 75.6\%. Thus, knowing a rule is not the same as using it over
time.

\begin{table}[t]
    \caption{Semantic accuracy on the 685-probe task audit
    (Qwen2.5-7B-Instruct), by probe subtask.}
    \label{tab:task-audit}
    \centering
    \small
    \begin{tabular}{@{}lr@{}}
        \toprule
        Subtask & Accuracy \\
        \midrule
        Rule recall                  & 97.4\%  \\
        One-stage payoff lookup      & 100.0\% \\
        Terminal scoring             & 53.3\%  \\
        State reconstruction         & 37.0\%  \\
        State transition             & 22.2\%  \\
        Expected-payoff calculation  & 16.7\%  \\
        \midrule
        Overall (all subtasks pooled) & 59.1\% \\
        \bottomrule
    \end{tabular}
\end{table}

This gap between local rule recall and reliable execution of the evolving
game state is the lens through which every subsequent behavioural result in
this section should be read: an agent can answer questions about the rules
correctly while still failing to track the state those rules act on. Full
probe-by-probe breakdowns are given in
Appendix~\ref{app:extended-results}.

\subsection{Strategic robustness under equivalent presentations}
\label{sec:results-sensitivity}

We next ask whether sampled behaviour is stable under manipulations that
leave the underlying game mechanics unchanged.

\paragraph{Disclosed arithmetic (clean diagnostic).} The canonical and
decision-card conditions each completed 30 races and 558 decisions with
zero parse failures; matched risk-by-repetition cells had identical
realised horizons. The decision card discloses the focal agent's immediate
payoff, progress, and private-risk consequences for each action profile
without revealing the opponent's move or the stopping horizon.
Table~\ref{tab:disclosed-arithmetic} compares the two conditions. Only
3.3\% of paired first-round decisions changed, so most divergence appeared
later in the race histories, after feedback had accumulated. These values
show that supplying current-round arithmetic changes sampled behaviour;
they do not by themselves identify an internal world model or a
confirmatory causal effect.

\begin{table}[t]
    \caption{Disclosed-arithmetic diagnostic: canonical vs.\ decision-card
    conditions (30 races, 558 decisions each, zero parse failures).}
    \label{tab:disclosed-arithmetic}
    \centering
    \small
    \begin{tabular}{@{}lrlr@{}}
        \toprule
        Condition & \Unsafe{} rate & 95\% CI & Mean payoff \\
        \midrule
        Canonical      & 52.0\% & 48.0--55.9\% & 42.77 \\
        Decision card  & 60.8\% & 51.2--67.9\% & 42.21 \\
        \bottomrule
    \end{tabular}
\end{table}

\paragraph{Opaque symbolic mapping (confounded, comprehension gate failed).}
A stronger and more concerning form of sensitivity appears when the game is
described through an opaque code rather than the words \Safe{}/\Unsafe{}.
The temperature-zero context pilot completed 768 live races (13,680
decisions) and a separate fixed-state replay of 1,536 context--mapping
cells, with no final parse failures. All paired non-control trajectories
agreed at round one; divergence appeared only after feedback, with the
largest live full-trajectory contrast against the abstract control at 34.0
percentage points and the largest fixed-state contrast at 16.7 points. When
the symbol $P$ denoted \Safe{}, six context contrasts diverged from the
abstract control; when $Q$ denoted \Safe{} instead, every corresponding
contrast was zero, even though the two mappings describe the identical
game. We flag this result rather than treat it as confirmatory evidence for
two reasons: the mapping was balanced by repetition parity rather than
fully crossed within seed blocks, and, more importantly, the associated
comprehension gate failed outright, with state-update accuracy at 12.5\%
and terminal-scoring accuracy at 17.2\%. We therefore report this as
representation-sensitive \emph{observed behaviour under fixed mechanics},
not as evidence that the model understood or optimised the race.

Taken together, these two diagnostics show that plausible-looking action
sequences can shift substantially under manipulations that a correct
world model would treat as irrelevant, which is precisely why the
comprehension check of \S\ref{sec:results-validity} is a prerequisite
for interpreting any of the behavioural results that follow.

\subsection{Game theory versus prompted race behaviour}
\label{sec:results-theory}

The evolutionary benchmark predicts a sharp change with private risk. At its
main reference setting, predicted Unsafe play is 99.2\%, 98.0\%, and 1.9\% in
the low-, medium-, and high-risk conditions. This pattern follows the shift
from AU to CAS and then CS described in \S\ref{sec:theory-benchmark}. The
temperature-zero Qwen context pilot does not reproduce this phase change. In
the technology framing it produces 0\% Unsafe play at all three risk levels,
while other narrative skins under the same mechanics range from 0\% to about
33\%.

This is a boundary rather than a failed equilibrium test. The evolutionary
model describes strategy frequencies under selection and mutation; the LLM
experiment samples prompted self-play decisions. Their units are different.
The comparison nevertheless gives a useful game-theoretic insight: holding the
payoff mechanism fixed does not guarantee the behavioural pattern predicted by
the reduced strategy model. For the audited checkpoint, task representation
can be more visible in observed play than the theoretical risk transition.

\subsection{Two-player strategic diversity: humans and LLMs}
\label{sec:results-diversity}

We compare model trajectories with the human benchmark of
\citet{falling_behind_unsafe} directly on the literal first-five-round
action-and-position histories, without first reducing them to hand-designed
summary statistics. The comparison includes 420 trajectories from the seven
tested models' two-player baseline (60 per model: GPT-5-nano, GPT-5.4-nano,
Gemini-3-Flash, Gemini-3.1-Flash-Lite, Gemini-3.5-Flash-Lite, Claude Opus 5,
and Claude Sonnet 5) and 340 complete human trajectories from the public
de-identified dataset (one incomplete human record excluded); pooled
$N=760$. Each player is represented by 15 raw features,
\[
(\mathrm{own}_{1..5},\ \mathrm{opp}_{1..5},\ \mathrm{gap}_{1..5}),
\]
own action, opponent action, and own-minus-opponent progress gap entering
each of rounds 1--5 -- the recorded state and actions themselves, rather
than derived quantities such as unsafe rate or retaliation.

Mean \Unsafe{} play over rounds 1--5 is lowest for GPT-5-nano (17\%),
followed by Claude Sonnet 5 (22\%), Claude Opus 5 (33\%), GPT-5.4-nano
(54\%), and humans (56\%); the Gemini models range from 73\%
(Gemini-3-Flash) to 83\% (Gemini-3.1-Flash-Lite). No single aggregate rate
characterises the tested models, and closeness to the human mean is, on its
own, uninformative about anything beyond that one number.

\begin{figure}[t]
    \centering
    \includegraphics[width=0.44\textwidth]{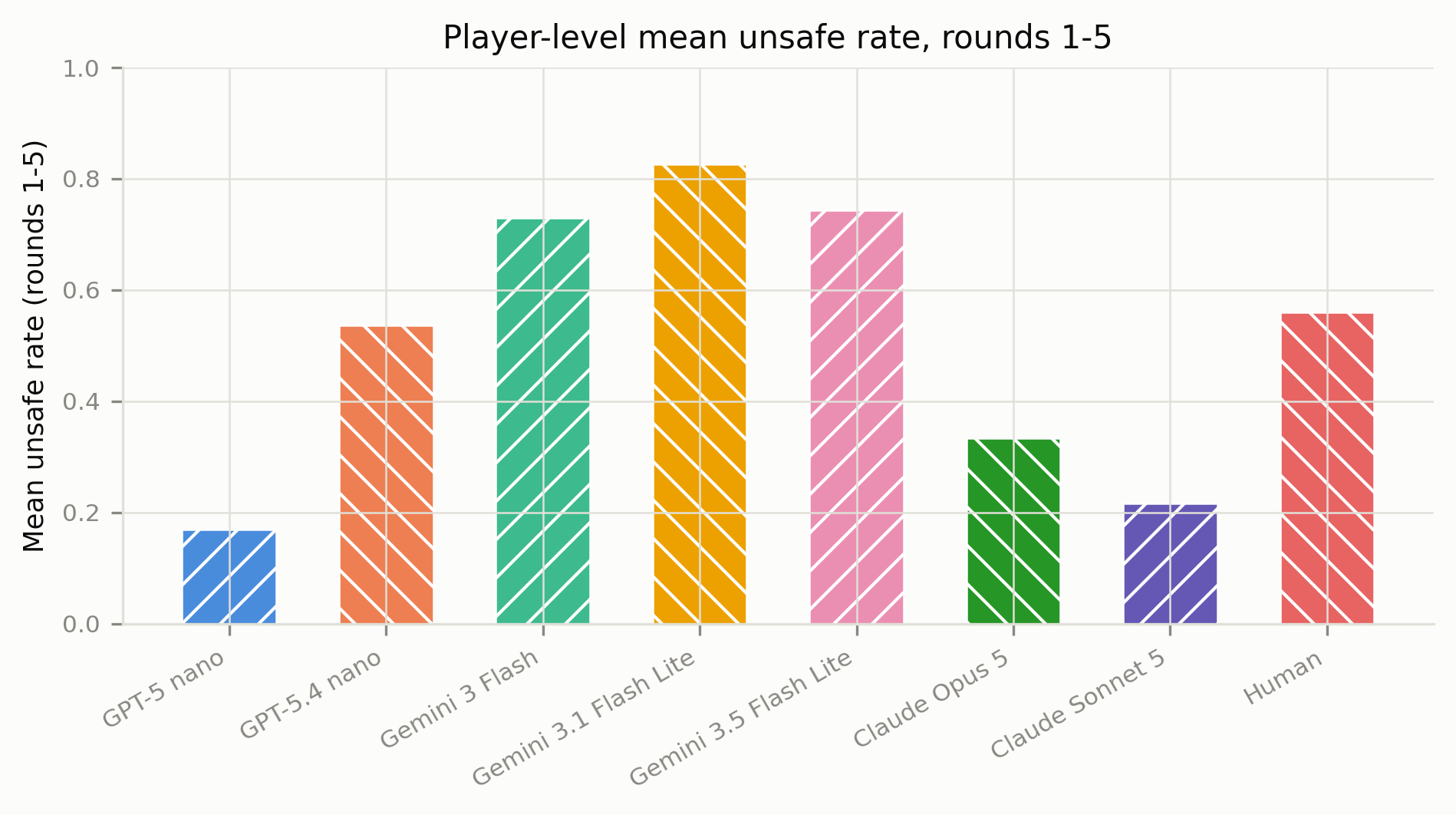}
    \caption{Player-level mean \Unsafe{} rate over rounds 1--5 per
    population, the numbers reported in the paragraph above.}
    \label{fig:unsafe-rate-by-group}
    \Description{Bar chart of mean Unsafe rate over rounds 1 to 5 for each
    population. GPT-5-nano is lowest at 17 percent, then Claude Sonnet 5
    at 22 percent, Claude Opus 5 at 33 percent, GPT-5.4-nano at 54 percent
    and humans at 56 percent. The three Gemini models are highest, ranging
    from 73 to 83 percent.}
\end{figure}

We next group similar trajectories without giving the algorithm predefined
strategy labels. HDBSCAN is a density-based clustering method: it finds dense
groups and leaves unusual cases unassigned. We call each resulting group a
behavioural \emph{archetype}, meaning a common action-and-position pattern,
not a mental type. With the fixed settings (minimum cluster size 15, minimum
samples 6, excess-of-mass selection), HDBSCAN identifies eleven archetypes
and leaves 98 of 760
trajectories unclustered (12.9\%). All eleven archetypes occur among the
human participants, whereas GPT-5-nano concentrates 95\% of its
trajectories into three predominantly-\Safe{} archetypes and 52\% of
GPT-5.4-nano's trajectories are not assigned to any dense archetype at all,
the highest unclustered share of any population. Claude Opus 5 and Claude
Sonnet 5 fall almost entirely (100\% and 93\%, respectively) into two
archetypes that both open with a mutual-\Safe{} first round.

\begin{figure}[t]
    \centering
    \includegraphics[width=0.48\textwidth]{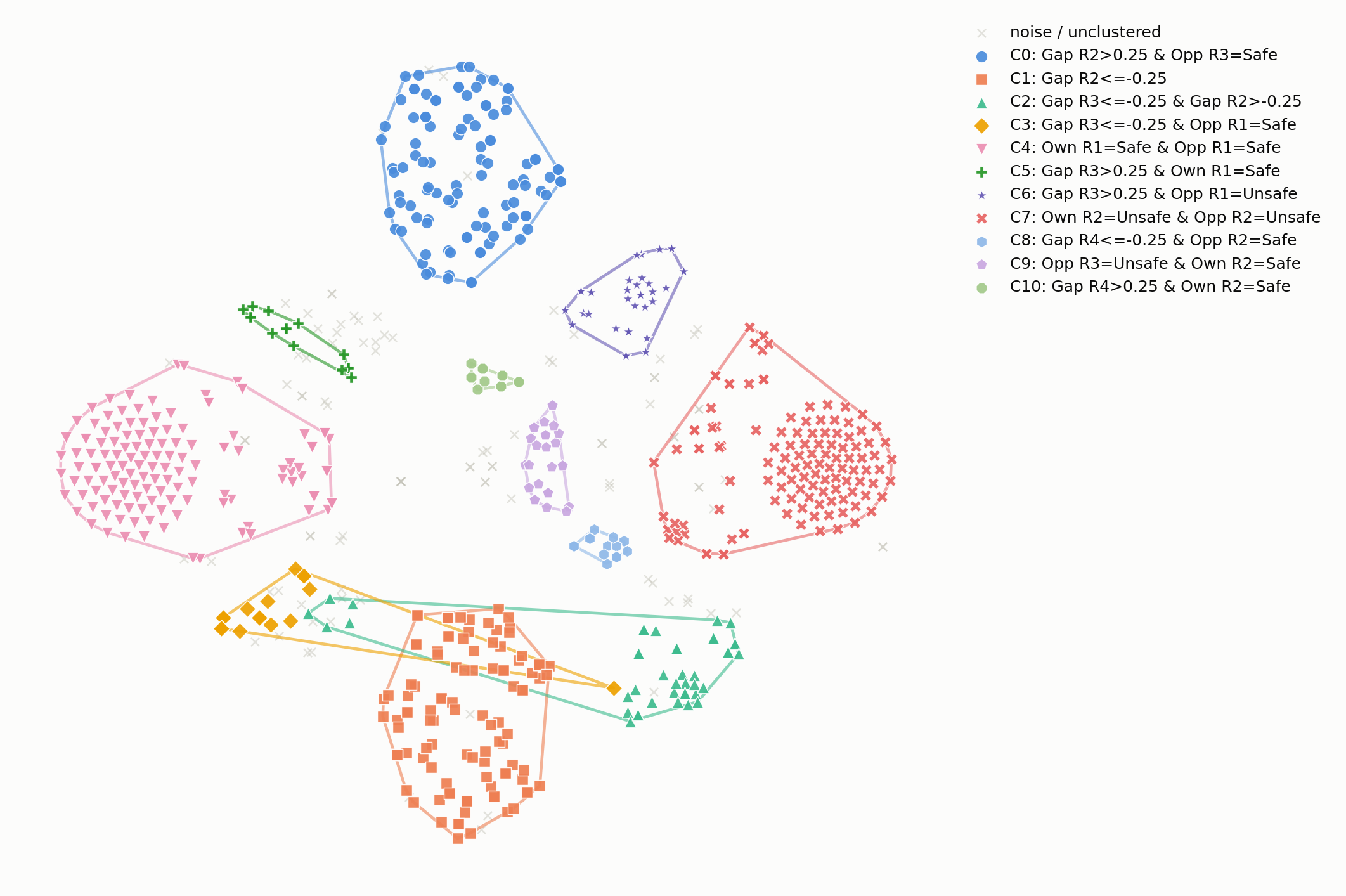}
    \caption{Composition of the eleven HDBSCAN-derived raw-action-and-
    position archetypes. The two largest are a mutual-first-round-\Safe{}
    archetype ($n=138$) and a mutual-round-2-\Unsafe{} archetype ($n=124$);
    98 of 760 trajectories (12.9\%) are unclustered.}
    \label{fig:hdbscan-composition}
    \Description{Composition of the eleven behavioural archetypes that
    HDBSCAN recovers from the raw five-round action-and-position
    sequences, showing how many trajectories of each population fall into
    each archetype. The two largest archetypes both begin with a mutual
    first move: one opens with mutual Safe play and contains 138
    trajectories, the other turns to mutual Unsafe play at round 2 and
    contains 124. Ninety-eight of the 760 trajectories are left unassigned
    to any dense archetype.}
\end{figure}

\begin{figure}[t]
    \centering
    \includegraphics[width=0.48\textwidth]{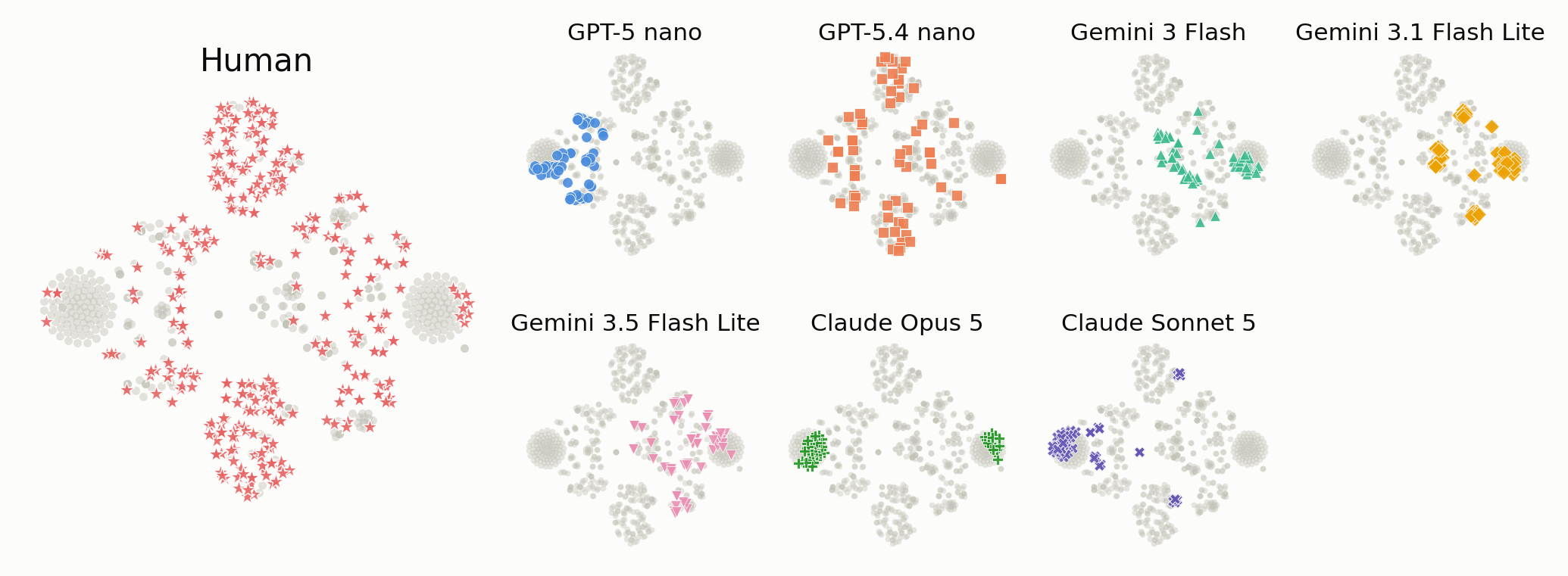}
    \caption{t-SNE visualisation of the pooled 15-dimensional raw
    action-and-position space. Each panel highlights one population while
    rendering the other populations in grey; the human sample is shown in
    the larger left panel, with the seven LLM checkpoints arranged smaller
    to its right.}
    \label{fig:human-tsne}
    \Description{Two-dimensional t-SNE embedding of the pooled
    fifteen-dimensional space of raw actions and progress gaps. The layout
    is a set of panels, one per population, in which the highlighted
    population is drawn in colour and every other population in grey. The
    human sample occupies the large panel on the left and the seven
    language-model checkpoints are arranged in smaller panels to its
    right. Human points are spread widely across the embedding, while each
    model concentrates in a narrower region: GPT-5-nano in the low-Unsafe
    area and Gemini 3.1 Flash Lite near the high-Unsafe area.}
\end{figure}

The embedding in Figure~\ref{fig:human-tsne} makes the difference between
behavioural \emph{level} and behavioural \emph{diversity} visible:
GPT-5-nano is concentrated in the low-\Unsafe{} region, Gemini-3.1-Flash-
Lite is concentrated near the high-\Unsafe{} region, and human trajectories
occupy a much broader portion of the observed space, overlapping several
model-specific regions.

\begin{figure}[t]
    \centering
    \includegraphics[width=0.48\textwidth]{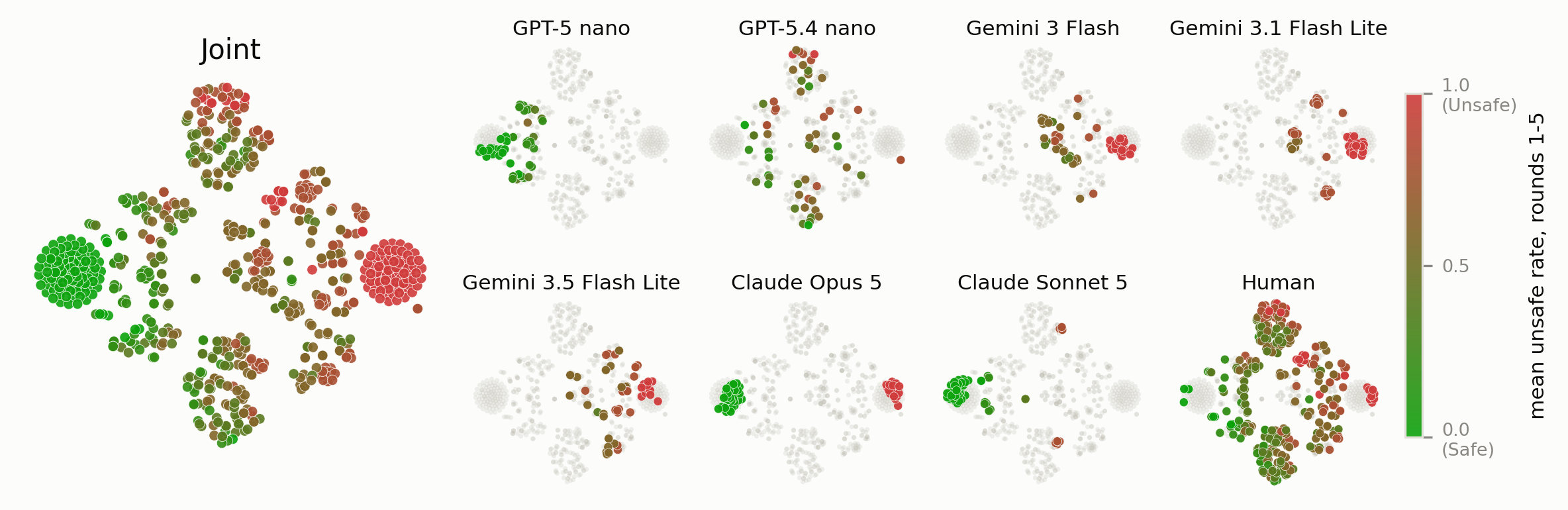}
    \caption{The same t-SNE coordinates as Figure~\ref{fig:human-tsne},
    recoloured by each player's own mean \Unsafe{} rate over rounds 1--5
    (green = \Safe{}, red = \Unsafe{}) instead of population identity.
    Left: all 760 trajectories pooled. Right: the same colouring broken out
    per population.}
    \label{fig:human-tsne-outcome}
    \Description{The same t-SNE coordinates as the previous figure,
    recoloured by each player's own mean Unsafe rate over rounds 1 to 5
    rather than by which population produced it, on a green-to-red scale
    where green is Safe and red is Unsafe. The left panel pools all 760
    trajectories and the right panel breaks the same colouring out per
    population. The embedding separates almost entirely by Unsafe rate: a
    solid green region, a solid red region, and a mixed band between them,
    with very few red points inside the green region or the reverse.}
\end{figure}

Recolouring the identical coordinates by outcome instead of identity
(Figure~\ref{fig:human-tsne-outcome}) shows the embedding separates almost
entirely by \Unsafe{} rate: a solid-green region, a solid-red region, and a
mixed band in between, with almost no red points inside the green region or
vice versa. The dense archetypes in Figure~\ref{fig:hdbscan-composition}
sit inside the solid-coloured ends of this gradient, while the more
diffuse, harder-to-cluster trajectories sit in the mixed middle band: the
raw-sequence archetypes HDBSCAN finds are not arbitrary with respect to
game outcome.

We next test whether five rounds contain enough information to identify the
population that produced a trajectory. A small decision tree reaches
$38.9\%\pm3.3$ accuracy under five-fold cross-validation. Random guessing
among eight balanced classes would give 12.5\%. Population identity is
therefore visible, but the groups still overlap. The tree correctly identifies
96.7\% of Gemini-3.1-Flash-Lite trajectories. For GPT-5-nano, this value is
63.3\%, and 31.7\% are instead labelled Claude Opus 5 because both often start
with mutual Safe play. Only 33.2\% of human trajectories are labelled human.
This is the lowest value among the populations. Human actions overlap with
several model patterns because the human sample is internally diverse, not
because it follows one middle policy.

\begin{figure}[t]
    \centering
    \includegraphics[width=0.48\textwidth]{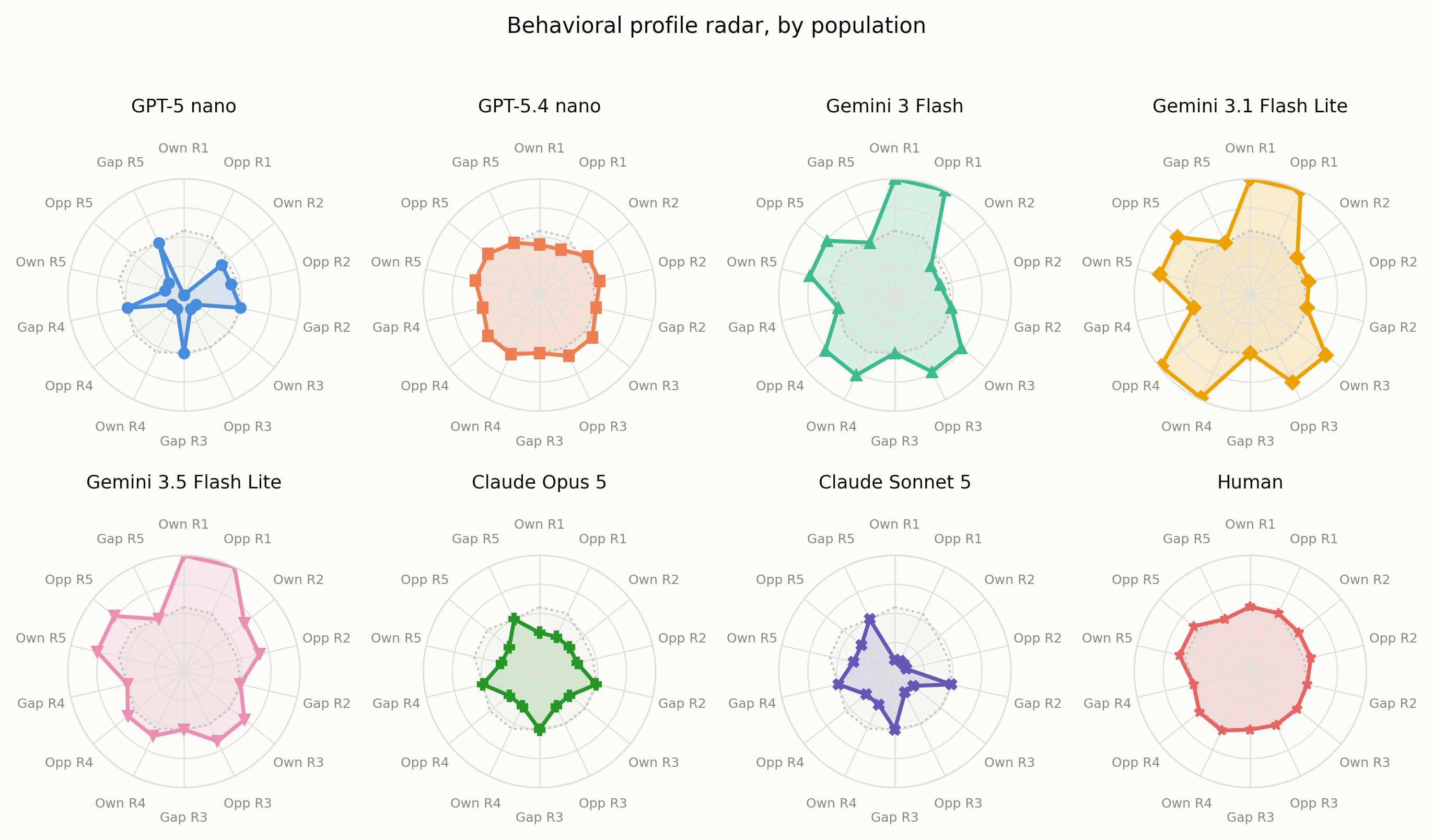}
    \caption{Mean round-by-round profile for each population, on the 14
    features shared across every trajectory (own and opponent action at
    rounds 1--5, plus progress gap entering rounds 2--5; the round-1 gap is
    omitted because every race starts at gap 0 by construction and is
    therefore uninformative). Solid lines show the population mean; the
    dotted line is the pooled mean across all eight populations.}
    \label{fig:human-radar}
    \Description{Small-multiple radar charts, one per population, over the
    fourteen features shared by every trajectory: own and opponent action
    at rounds 1 to 5, plus the progress gap entering rounds 2 to 5. Each
    panel draws that population's mean profile as a solid line against the
    pooled mean of all eight populations as a dotted reference line. The
    human profile sits close to the pooled mean on nearly every axis.
    GPT-5-nano and Claude Sonnet 5 are displaced toward Safe, the two
    Gemini Flash Lite checkpoints toward Unsafe, and GPT-5.4-nano, Claude
    Opus 5 and Gemini 3 Flash fall between those extremes.}
\end{figure}

The round-by-round profile in Figure~\ref{fig:human-radar} gives a
complementary view to the aggregate rate: the human profile lies close to
the pooled eight-population mean on nearly every own-, opponent-, and
gap-action axis, whereas GPT-5-nano and Claude Sonnet 5 are displaced
toward \Safe{}, the two Gemini Flash-Lite checkpoints are displaced toward
\Unsafe{}, and GPT-5.4-nano, Claude Opus 5, and Gemini-3-Flash fall between
these extremes. Humans therefore do not occupy a single isolated
behavioural corner: their aggregate profile is comparatively central, while
their individual trajectories span the regions occupied by several
different models.

This heterogeneity is not merely a visual impression. Fitting three nested
logistic models to the pooled neutral-baseline pilot (risk-only,
LLM-identity-only, and LLM-identity-by-risk interaction) shows that
allowing the risk-response profile to vary by LLM improves fit
substantially beyond LLM-specific intercepts alone,
\[
    \chi^2(10)=354.7,\qquad p\approx4\times10^{-70},
\]
with several risk-by-LLM cells approaching complete separation (a
convergence warning in the interaction model means the exact statistic
should be read as approximate). The fit covers the five-model neutral
baseline only; the two Claude models are not in it. Neither caveat changes
the conclusion: no single risk-response curve represents all five, so
cross-model differences are differences in \emph{shape}, not only in
level.

\begin{figure}[t]
    \centering
    \includegraphics[width=\columnwidth]{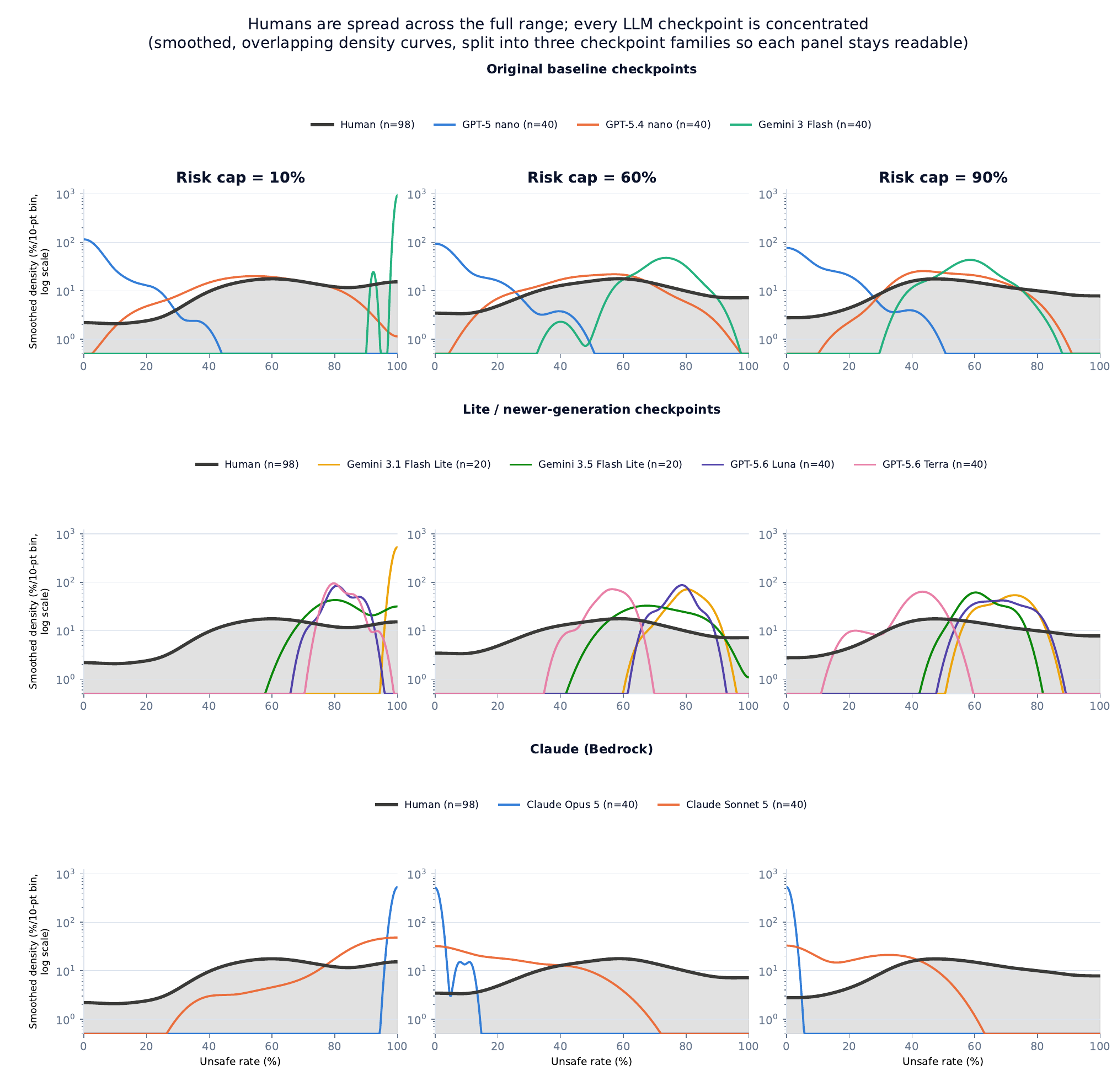}
    \caption{Smoothed distribution of player-level \Unsafe{} rates by risk
    cap (columns) and model family (rows), with the human benchmark
    repeated in every panel as the shaded reference. Density is log-scaled:
    model peaks span close to three orders of magnitude, and a linear axis
    would flatten all but the tallest. The middle row includes GPT-5.6 Luna
    and Terra, outside the seven-model roster.}
    \label{fig:human-llm-distribution}
    \Description{A three-by-three grid of smoothed density curves of
    player-level Unsafe rate. Columns are the three risk caps, 10, 60 and
    90 percent. Rows are three model families: the original baseline
    checkpoints, the Flash Lite and newer-generation checkpoints, and the
    two Claude checkpoints. The human benchmark is repeated in every panel
    as a shaded grey reference curve. The horizontal axis is Unsafe rate
    from 0 to 100 percent and the vertical axis is a log-scaled density.
    In every panel the human curve is low and broad across almost the
    whole range, while each model curve rises to a tall narrow peak at its
    own characteristic Unsafe rate.}
\end{figure}

This distributional picture is visible directly in the raw \Unsafe{}-rate
data as well (Figure~\ref{fig:human-llm-distribution}): human
participant-level \Unsafe{} rates spread across nearly the entire
available range at each risk level, whereas each tested model
occupies a comparatively narrow, model-specific band: GPT-5-nano
concentrated near the low-\Unsafe{} end, the Gemini models toward the
high-\Unsafe{} end, and GPT-5.4-nano in a more intermediate, diffuse
region. Consequently, similarity between one model's aggregate
\Unsafe{} rate and the human mean does not imply that the model
reproduces the underlying human distribution.

A more limited correspondence appears in the relationship between race
outcome and behaviour: human winners play \Unsafe{} 16.0 percentage points
more often than human losers, versus an approximately 20-point pooled
descriptive difference among the tested models. This agreement is
confined to one marginal association and does not extend to the full
dynamic structure: the human estimates show a precise positive association
with the opponent's previous action and a negative association with
relative race position, whereas several model estimates are weak,
reversed in sign, or not reliably estimable because their action
distributions sit near floor or ceiling. Differences in sign, magnitude,
and precision across models confirm that agreement in an aggregate
action rate does not imply agreement in the underlying behavioural
dynamics.

Finally, we ask whether models reproduce the \emph{diversity} of human
strategies, not just an aggregate rate. This uses the full human sample of
341 participants rather than the 340-trajectory subset used in the
HDBSCAN comparison above, since none of the five
descriptors below requires a complete five-round export. Using five
trajectory-level descriptors computed from these 341 human participants
(overall \Unsafe{} rate, opponent reciprocity, position sensitivity,
own-action autocorrelation, first-round action), $k$-means at $k=4$ yields
four
reference archetypes (\emph{cautious-starter}, \emph{aggressive-starter/%
reciprocator}, \emph{reciprocal catch-up}, and \emph{persister}) with a
modest silhouette score (0.24--0.32 across $k\in\{3,4,5\}$), so this
partition should be read as a noisy exploratory organisation of the human
data rather than a definitive taxonomy. Projecting each model's
trajectories into this human-only reference space, using human means and
standard deviations throughout, gives
Table~\ref{tab:archetype-coverage}.

\begin{table}[t]
  \caption{Behavioural-archetype coverage: share of each population's
  neutral-lane player-races assigned to each human-fit archetype.
  Human row is the reference clustering ($n{=}341$); each LLM row
  projects that checkpoint's own no-persona player-races
  ($n{=}60$--$120$, see Table~\ref{tab:evidence-strata}) into the same
  space. This table draws on the full nine-checkpoint cross-model pilot
  and therefore includes GPT-5.6 Luna and Terra, which are not part of
  the seven-model HDBSCAN/t-SNE/decision-tree comparison earlier in
  this subsection.}
  \label{tab:archetype-coverage}
  \centering
  \footnotesize
  \setlength{\tabcolsep}{3pt}
  \begin{tabular}{@{}lrrrr@{}}
    \toprule
    Population & Cautious & Aggr./recip. & Catch-up & Persister \\
    \midrule
    Human (reference)      & 39.0\% & 49.9\% & 5.6\%  & 5.6\% \\
    GPT-5-nano              & 99.2\% & 0\%    & 0\%    & 0.8\% \\
    GPT-5.4-nano            & 51.7\% & 33.3\% & 3.3\%  & 11.7\% \\
    Gemini-3-Flash          & 1.7\%  & 78.3\% & 20.0\% & 0\% \\
    Gemini-3.1-Flash-Lite   & 0\%    & 83.3\% & 16.7\% & 0\% \\
    Gemini-3.5-Flash-Lite   & 0\%    & 76.7\% & 23.3\% & 0\% \\
    GPT-5.6 Luna            & 3.3\%  & 72.5\% & 24.2\% & 0\% \\
    GPT-5.6 Terra           & 14.2\% & 59.2\% & 26.7\% & 0\% \\
    Claude Opus 5           & 66.7\% & 33.3\% & 0\%    & 0\% \\
    Claude Sonnet 5         & 77.5\% & 20.8\% & 1.7\%  & 0\% \\
    \bottomrule
  \end{tabular}
\end{table}

GPT-5-nano is effectively a point mass in the cautious-starter archetype
(99.2\%): its low aggregate rate reflects a highly compressed behavioural
distribution, not broad coverage of cautious human behaviour. The three
Gemini models are similarly compressed into the aggressive-starter and
catch-up archetypes and contribute essentially no cautious or persister
trajectories, and GPT-5.6 Luna and Terra show the same pattern. Claude
Opus 5 and Claude Sonnet 5 also
concentrate in the cautious-starter archetype (66.7\% and 77.5\%,
respectively), but, unlike GPT-5-nano, split the remainder with the
aggressive-starter/reciprocator archetype (33.3\% and 20.8\%) rather than
collapsing to a near-total point mass. GPT-5.4-nano is the only model
represented in all four
archetypes, consistent with its diffuse behavioural profile elsewhere in
this section, but its cluster proportions still differ from the human
distribution: proximity to a centroid is a resemblance measure, not
evidence of psychological or mechanistic equivalence. Human-like behaviour
is therefore multidimensional: a model can resemble one human
archetype while failing to reproduce the diversity observed across human
participants.

\subsection{Observable drivers of \Unsafe{} play}
\label{sec:results-shap}

Aggregate action rates do not show how agents respond to a changing race.
We therefore test how five pre-decision variables predict Unsafe play: the
player's previous action, the opponent's previous action, progress gap,
assigned private risk, and round number. These are predictive associations,
not causal effects or evidence about internal reasoning. The full random-
forest and SHAP analysis is reported in Appendix~\ref{app:predictive-structure}.

The human data are organised mainly by the opponent's previous action, which
accounts for 48\% of the fitted model's total SHAP magnitude. The tested LLMs
do not share one pattern. Progress gap is largest for GPT-5-nano (44\%), the
assigned risk condition is largest for two Gemini checkpoints (35--40\%), and
opponent history is largest for Claude Sonnet 5 (51\%). GPT-5.4-nano has no
single strong predictor. These results support a game-theoretic point: two
populations can have similar Unsafe rates while following different response
rules. Appendix~\ref{app:predictive-structure} reports model fit, confounding,
and floor/ceiling limits.

\subsection{Multi-agent dynamics: position in \texorpdfstring{$N$}{N}-player races}
\label{sec:results-position}

The two-player analysis uses the opponent's previous action and the exact
progress gap. In races with more players, we use a simpler question: does an
agent act differently when it is a Leader, in the Middle, or a Trailer? We
compare these positions within four persona bands. Baseline adds no persona
sentence; Low combines R1--R2; Mid combines R3--R4; and High combines R5--R6.
A Leader has the highest progress or is tied for it before the current round.
A Trailer is strictly last. Middle covers every other position. These labels
describe observed states; the experiment did not randomly assign rank.

\begin{figure}[t]
    \centering
    \includegraphics[width=0.48\textwidth]{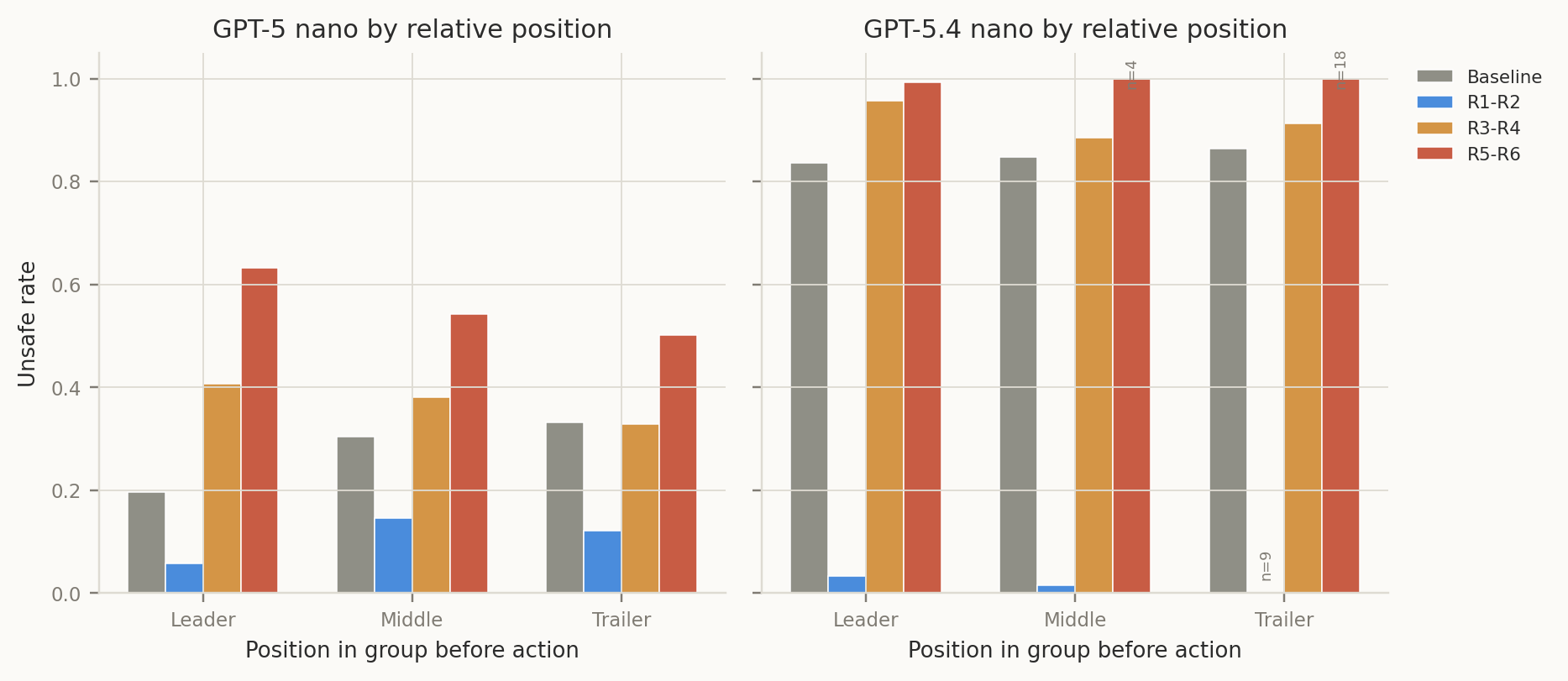}
    \caption{\Unsafe{} rate by relative position (Leader / Middle /
    Trailer) within each assigned persona band, for GPT-5-nano (left)
    and GPT-5.4-nano (right). Bars marked $^\dagger$ in
    Tables~\ref{tab:position-gpt5nano}--\ref{tab:position-gpt54nano} have
    $n<20$ and should be read as noisy pilot estimates rather than stable
    rates.}
    \label{fig:position-by-risk}
    \Description{Grouped bar chart of Unsafe rate by a company's relative
    position, Leader, Middle or Trailer, within each assigned persona
    band, for GPT-5-nano on the left and GPT-5.4-nano on the right. For
    GPT-5-nano the Trailer bar is taller than the Leader bar in the
    Baseline and Low bands and shorter in the Mid and High bands, so the
    ordering reverses. For GPT-5.4-nano the three position bars are close
    to equal within every band, and the bands themselves move from near
    the floor at Low to near the ceiling at Mid and High.}
\end{figure}

\begin{table}[t]
    \caption{\Unsafe{} rate by persona band and relative position,
    GPT-5-nano. $n$ is the number of decisions in the cell; CI is the
    95\% confidence interval on the unsafe rate.}
    \label{tab:position-gpt5nano}
    \centering
    \small
    \begin{tabular}{@{}llrrl@{}}
        \toprule
        Persona band & Position & $n$ & \Unsafe{} rate & 95\% CI \\
        \midrule
        Baseline      & Leader  & 1460 & 19.7\% & 17.7--21.8\% \\
        Baseline      & Middle  & 1065 & 30.4\% & 27.7--33.3\% \\
        Baseline      & Trailer & 175  & 33.1\% & 26.6--40.4\% \\
        Low (R1--R2)  & Leader  & 1094 & 5.9\%  & 4.6--7.4\%   \\
        Low (R1--R2)  & Middle  & 198  & 14.6\% & 10.4--20.2\% \\
        Low (R1--R2)  & Trailer & 58   & 12.1\% & 6.0--22.9\%  \\
        Mid (R3--R4)  & Leader  & 649  & 40.7\% & 37.0--44.5\% \\
        Mid (R3--R4)  & Middle  & 451  & 38.1\% & 33.8--42.7\% \\
        Mid (R3--R4)  & Trailer & 250  & 32.8\% & 27.3--38.8\% \\
        High (R5--R6) & Leader  & 702  & 63.2\% & 59.6--66.7\% \\
        High (R5--R6) & Middle  & 339  & 54.3\% & 49.0--59.5\% \\
        High (R5--R6) & Trailer & 309  & 50.2\% & 44.6--55.7\% \\
        \bottomrule
    \end{tabular}
\end{table}

\begin{table}[t]
    \caption{\Unsafe{} rate by persona band and relative position,
    GPT-5.4-nano. Cells marked $^\dagger$ have $n<20$.}
    \label{tab:position-gpt54nano}
    \centering
    \small
    \begin{tabular}{@{}llrrl@{}}
        \toprule
        Persona band & Position & $n$ & \Unsafe{} rate & 95\% CI \\
        \midrule
        Baseline      & Leader  & 1649 & 83.6\%  & 81.8--85.3\%  \\
        Baseline      & Middle  & 703  & 84.8\%  & 81.9--87.2\%  \\
        Baseline      & Trailer & 348  & 86.5\%  & 82.5--89.7\%  \\
        Low (R1--R2)  & Leader  & 1073 & 3.3\%   & 2.4--4.5\%    \\
        Low (R1--R2)  & Middle  & 268  & 1.5\%   & 0.6--3.8\%    \\
        Low (R1--R2)  & Trailer & 9$^\dagger$    & 0.0\%   & 0.0--29.9\%   \\
        Mid (R3--R4)  & Leader  & 1164 & 95.7\%  & 94.4--96.7\%  \\
        Mid (R3--R4)  & Middle  & 70   & 88.6\%  & 79.0--94.1\%  \\
        Mid (R3--R4)  & Trailer & 116  & 91.4\%  & 84.9--95.3\%  \\
        High (R5--R6) & Leader  & 1328 & 99.3\%  & 98.7--99.6\%  \\
        High (R5--R6) & Middle  & 4$^\dagger$    & 100.0\% & 51.0--100.0\% \\
        High (R5--R6) & Trailer & 18$^\dagger$   & 100.0\% & 82.4--100.0\% \\
        \bottomrule
    \end{tabular}
\end{table}

For GPT-5-nano, position matters, but its effect flips sign across persona
bands (Table~\ref{tab:position-gpt5nano}). At Baseline and Low, \Unsafe{}
rate rises as rank falls, the qualitative direction predicted by the human
``falling behind'' account of \citet{falling_behind_unsafe}: the company
that is behind takes on more risk. At Mid and High, the ordering reverses,
with Leaders now \emph{more} \Unsafe{} than Trailers. We do not read this
reversal as evidence that being in the lead causes more risk-taking under
more risk-seeking framing. Because rank is measured immediately before the
action rather than assigned independently of it, a company's own tendency
to play \Unsafe{} and its current rank are entangled: a company that has
already played \Unsafe{} more often is, mechanically, more likely to have
accumulated the extra progress that makes it the Leader
($\sigma(\Unsafe{})=1.5$ vs.\ $\sigma(\Safe{})=1.0$). Under Mid/High
framing, where the model plays \Unsafe{} often enough for this composition
effect to dominate, the Leader group is disproportionately populated by
already-\Unsafe{}-leaning trajectories, which would produce exactly this
reversal without any causal effect of \emph{being ahead} on the next
choice. Distinguishing this selection story from a genuine effect of rank
would require conditioning on, or randomising, prior own-action history
within rank; we flag this as a concrete next step rather than resolve it
here.

GPT-5.4-nano shows a starkly different pattern: the assigned persona band
dominates and rank contributes little on top of it
(Table~\ref{tab:position-gpt54nano}). \Unsafe{} play is uniformly low at
Low and uniformly near-ceiling at Mid and High, with Leader, Middle, and
Trailer tracking each other closely within each band, including at
Baseline itself, where all three ranks already sit in a similar range
before any persona is applied. This is consistent with the
model's SHAP profile (Table~\ref{tab:feature-importance}), where
progress gap contributes only 33\% of predictive importance against a more
evenly spread set of features, so relative standing is not the dominant
driver of its \Unsafe{} choices. The three
cells marked $^\dagger$ in Table~\ref{tab:position-gpt54nano} have
confidence intervals too wide for their point estimates to be treated as
stable; they arise because so few races reach a Trailer or Middle rank at
all once GPT-5.4-nano is pushed toward near-uniform \Unsafe{} play under
Mid/High framing, which itself thins out the population of races with any
surviving rank diversity to measure.

Read together, the two models show that ``does position matter'' has
no single answer even within one model family: GPT-5-nano displays a
falling-behind-consistent pattern at Baseline/Low framing that reverses
under Mid/High framing in a way better explained by selection than by a
causal effect of rank, while GPT-5.4-nano's behaviour is essentially
saturated by the assigned persona band regardless of rank. In both cases,
the assigned persona, not relative standing, is the primary
explanatory variable; position is, at most, a secondary modulator whose
apparent effect depends on which framing a company was given. The bands
here are the prompt-embedded risk \emph{persona} examined next in
\S\ref{sec:results-persona}, not the mechanism's private-risk parameter
$p_r^{\max}$ (Eq.~\ref{eq:risk}), and each band pools over $p_r^{\max}$
rather than fixing it. The distinction matters: the persona is an
instruction re-assigned per race, $p_r^{\max}$ a property of the game. This
section therefore shows rank effects modulated by persona; separating that
from the mechanical risk treatment needs a design crossing the two, which
this pilot lacks. The broader
$N\in\{3,4,5\}$ self-play scope (180 pilot races, 6,120 decisions;
aggregate \Unsafe{} rates of 16.2\%, 26.8\%, 21.9\% for GPT-5-nano and
83.7\%, 87.0\%, 84.0\% for GPT-5.4-nano at $N=3,4,5$) is reported in
Appendix~\ref{app:extended-results}; because changing $N$ also changes the
stage-payoff table, opponent count, and prompt length, we do not treat
that comparison as an isolated group-size effect.

\subsection{Measured human risk and prompted LLM personas}
\label{sec:results-persona}

Human risk preference and an LLM risk persona are different variables. Human
participants completed an incentivised risk task before the race. Their score
has almost no direct relation with later Unsafe play ($r=-0.015$, $p=0.79$,
$n=341$). By contrast, an LLM persona is an instruction placed inside every
race prompt. Across models with a six-level sweep, moving from the lowest to
the highest risk persona raises Unsafe play by 50.5 to 98.3 percentage points.

The contrast supports a behavioural, not psychological, interpretation. The
human measure describes a person before play; the LLM label actively changes
the decision context. A risk persona therefore acts as a strong policy prompt,
not as evidence that the model has a stable risk preference. Full curves,
model-level estimates, and scope limits are reported in
Appendix~\ref{app:persona-detail}.

\section{Discussion and limitations}
\label{sec:discussion}

The main lesson is not that an audited model ``understands the game.'' The
evidence supports a narrower claim. The model can recall rules and find local
payoffs, yet still make regular errors when it updates the state or calculates
expected payoff. Its action sequences also change when verified arithmetic or
an equivalent response representation is provided. Reports of LLM behaviour
in repeated games should therefore show comprehension accuracy and parser
health together with action rates.

The human experiment gives us an external behavioural reference. It is not a
target that an LLM should reproduce. When an LLM matches a human association,
the two populations behave similarly on that measured part of this task. This
does not show that they use the same reasoning or psychology. A mismatch also
does not make the human result invalid. Model, persona, and multiplayer
comparisons have the same boundary: each applies only to the tested endpoint,
prompt, decoding setting, provider route, and run period.

Four limitations shape what we can claim. First, the context-to-symbol mapping
changes with repetition parity, so mapping and repetition are not fully
separated. Second, the related comprehension gate failed. We can describe the
result as prompt-sensitive output, but not as informed game optimisation.
Third, fixed-state replay measures one decision per saved state, while live
play measures dependent decisions along a changing trajectory; their effect
sizes are not directly interchangeable. Fourth, the multiplayer pilots are
small, come from mixed sources, and do not include a matched two-player
condition. Changing the number of players also changes the payoff table,
opponent count, and prompt length. We therefore cannot identify a pure group-
size effect. More broadly, this game models only a private speed--safety trade-
off. It does not model the full structure of frontier AI labs, collective
harm, regulation, communication, or uncertainty about real capabilities.



\section{Conclusion}

Strategic safety behaviour in these AI races is not a fixed property of an LLM.
In two-player races, models differ in their responses to risk, opponents, and
relative progress. Across three to five agents, position patterns vary by model
and persona and are not monotone in group size. Yet rule recall does not ensure
state tracking, and equivalent presentations can change later play. Multi-agent
AI-race studies should therefore report dynamic response rules and validity,
not only aggregate Unsafe rates. These findings remain exploratory until
matched confirmatory runs pass the same gates.

\section*{Acknowledgements}
T.A.H. acknowledges travel and accommodation support from the Ho Chi Minh City University of Technology (HCMUT), VNUHCM (Adjunct Professorship scheme HCMUT\text{-}VNUHCM). L.H.T acknowledges the Ho Chi Minh City University of Technology (HCMUT), VNUHCM for supporting this study.
 TAH is supported by EPSRC (grant EP/Y00857X/1).
 
\appendix
\section{Extended results and robustness checks}
\label{app:extended-results}

\providecommand{\Safe}{\textsc{Safe}}
\providecommand{\Unsafe}{\textsc{Unsafe}}
\providecommand{\todo}[1]{\textbf{[TODO: #1]}}

This appendix collects the material referenced from
Section~\ref{sec:results} as supporting rather than extending the main
claims: the full task-validity probe breakdown
(\S\ref{app:task-validity-full}), the human-benchmark replication check
(\S\ref{app:human-benchmark-replication}), the persona comparison
(\S\ref{app:persona-detail}), the predictive analysis
(\S\ref{app:predictive-structure}), the HDBSCAN sensitivity check
(\S\ref{app:hdbscan-sweep}), and the full two-player and $N$-player
descriptive results (\S\ref{app:twoplayer-baseline},
\S\ref{app:nplayer-scope}).

\subsection{Task-validity probe battery: full breakdown}
\label{app:task-validity-full}

Table~\ref{tab:task-validity-full} reports the full breakdown behind the
single semantic-accuracy figure (59.1\%) quoted in
\S\ref{sec:results-validity}. The admitted task audit used
Qwen2.5-7B-Instruct under protocol \texttt{ai-race-game-understanding-v2}:
41 atomic probes spanning rule recall, one-stage payoffs, state
reconstruction, state transition, terminal scoring, and expected payoff,
with 685 planned fixed-seed probe outputs, all retained. Numeric probes
varied direct wording, paraphrase, and calculator disclosure; categorical
probes additionally reversed answer order.

\begin{table}[h]
    \caption{Task-validity probe accuracy by category, Qwen2.5-7B-Instruct,
    protocol \texttt{ai-race-game-understanding-v2} ($n=685$ retained
    probe outputs).}
    \label{tab:task-validity-full}
    \centering
    \small
    \begin{tabular}{@{}lr@{}}
        \toprule
        Probe category & Accuracy \\
        \midrule
        Rule recall                        & 97.4\% \\
        One-stage payoff lookup             & 100.0\% \\
        State reconstruction                & 37.0\% \\
        State transition                    & 22.2\% \\
        Terminal scoring                    & 53.3\% \\
        Expected-payoff calculation         & 16.7\% \\
        \midrule
        \emph{Overall semantic accuracy}    & \emph{59.1\%} \\
        \emph{Strict output-format compliance} & \emph{32.1\%} \\
        \bottomrule
    \end{tabular}
\end{table}

Two further contrasts qualify these category-level numbers. First, semantic
accuracy without a disclosed calculator result was 52.1\%, versus 75.6\%
when the verified result was shown -- the same clean-diagnostic comparison
discussed behaviourally in \S\ref{sec:results-sensitivity}, here measured
at the level of task comprehension rather than sampled action. Second,
answer-order reversal did not change correctness on the tested categorical
probes, whereas paraphrasing the same question changed correctness on some
state-reconstruction and terminal-scoring items; this separates
\emph{repeatability under a fixed decoding contract} from
\emph{robustness of the underlying measurement}, and is part of why we
treat state-level probes as less reliable than rule-recall probes even
before looking at any behavioural result.

At the level of individual paired probe items, none of the 7 order-reversal
pairs (spanning rule recall and terminal scoring) flipped either the raw
parsed answer or scored correctness (0/7). Of the 41 paraphrase pairs, 12
flipped the raw parsed answer, but only 3 changed scored correctness:
\texttt{state\_own\_risk} (state reconstruction), and
\texttt{terminal\_opp\_final} and \texttt{terminal\_winner\_setback} (both
terminal scoring). The remaining 9 answer-only flips left correctness
unchanged because both paraphrase variants were already scored the same
way (both correct or both incorrect). This is the exact pairwise count
behind the qualitative claim above: paraphrase-induced correctness changes
concentrate in state- and terminal-level items, not rule-recall items.

\subsection{Human-benchmark replication check}
\label{app:human-benchmark-replication}

Before comparing model-generated behaviour with the human benchmark of
\citet{falling_behind_unsafe} in \S\ref{sec:results-diversity}, we first
validated our reconstruction of that benchmark by refitting its published
dynamic specification on the same public de-identified data. The
reconstructed estimation sample contains 2{,}888 round-level observations,
172 race-pair clusters, and 338 participants. Table~\ref{tab:human-replication} compares the refit coefficients against the values reported in the
original study.

\begin{table}[h]
    \caption{Replication of \citet{falling_behind_unsafe}'s dynamic
    specification on the public de-identified human data. ``Reconstructed''
    is our refit; ``Original'' is the coefficient reported in the source
    study.}
    \label{tab:human-replication}
    \centering
    \small
    \begin{tabular}{@{}lrr@{}}
        \toprule
        Term & Reconstructed & Original \\
        \midrule
        Opponent's previous action & 0.606  & 0.607  \\
        Player's progress gap      & $-0.295$ & $-0.296$ \\
        First-round action         & 0.217  & 0.217  \\
        Player's own previous action & $-0.195$ & $-0.193$ \\
        \bottomrule
    \end{tabular}
\end{table}

\subsection{Claim-scoped citation audit}
\label{app:citation-audit}

Table~\ref{tab:citation-audit} records both the supported use of each cited
study and the nearest excluded overclaim. These evidence boundaries do not
turn the diagnostic pilots in this paper into confirmatory results.

\begin{table*}[t]
\caption{Source-level audit for citations added to the prompt-sensitivity
and behavioural-validity argument.}
\label{tab:citation-audit}
\small
\begin{tabular}{@{}p{0.16\textwidth}p{0.38\textwidth}p{0.38\textwidth}@{}}
\toprule
Source & Supported use & Excluded overclaim \\
\midrule
Sclar et al.~\citep{sclarPromptFormatting2024} &
Meaning-preserving few-shot formatting changes can yield large,
model-dependent performance spreads. &
Not evidence for semantic paraphrase effects or a universal effect size. \\
Pezeshkpour and Hruschka~\citep{pezeshkpourOptionOrder2024} &
Reordering multiple-choice answer options can change predictions. &
The proposed uncertainty-plus-placement mechanism is not causally
identified. \\
Wei et al.~\citep{weiSelectionBias2024} &
Order and response-token identity are distinct selection-bias targets. &
Not proof that every task or model exhibits the same bias. \\
Salinas and Morstatter~\citep{salinasButterfly2024} &
Controlled output-format and atomic whitespace variants can alter answers. &
Not a claim that every added space changes behaviour. \\
Zheng et al.~\citep{zhengPersona2024} &
Across their factual-question setup, average persona benefit was absent or
slightly negative and reliable persona selection remained difficult. &
Not a universal null effect for persona-driven agent tasks. \\
Aher et al.~\citep{using_llm_to_simulate} &
Three of four behavioural findings were qualitatively reproduced, while the
fourth showed a hyper-accuracy distortion. &
Not evidence that LLM samples are interchangeable with human populations. \\
Akata et al.~\citep{playing_repeated_games_with_llms} &
Prediction before action improved coordination and scores in repeated games. &
Not a general claim that an observed action reveals a correct opponent
belief. \\
Herr et al.~\citep{herrStrategicBias2024} &
Action-label order and payoff reassignment changed choices in two canonical
two-player games. &
Not evidence that every strategically equivalent transformation has the same
effect. \\
Robinson and Burden~\citep{robinsonBurdenFraming2025} &
Procedurally varied vignettes produced substantial contextual variability
under one underlying Prisoner's Dilemma structure. &
Not a repeated-game result and not an estimate for the present race. \\
Wu et al.~\citep{wuCounterfactual2024} &
Performance declined consistently across 11 tasks when default mappings were
replaced by specified counterfactual alternatives. &
Not proof that the tested models lacked all abstract reasoning ability. \\
del Rio-Chanona et al.~\citep{delRioChanonaMarkets2025} &
LLM markets recovered broad feedback-condition dynamics but displayed less
strategy heterogeneity than human participants. &
Not evidence that LLM agents and human participants are exchangeable. \\
\bottomrule
\end{tabular}
\end{table*}

The reconstruction reproduces the original coefficients almost exactly
(largest discrepancy 0.002), which is the basis for treating the human-side
comparisons in \S\ref{sec:results-diversity} and
\S\ref{sec:results-persona} as resting on a correctly processed benchmark
rather than on a reimplementation error.

\subsection{Risk preference and assigned personas}
\label{app:persona-detail}

We finally distinguish three constructs that should not be conflated: the
human participants' pre-game elicited risk preference, the mechanism's
assigned private-risk treatment, and the narrative risk persona supplied to
an LLM in its prompt. Human participants completed an incentivised
Eckel--Grossman gamble elicitation (0 = most risk-averse, 5 = least
risk-averse) before playing the two-player race. This elicited measure has
essentially no direct association with subsequent \Unsafe{} rate,
\[
    \text{slope}=-0.004,\qquad r=-0.015,\qquad p=0.79,\qquad n=341,
\]
with the difference in mean \Unsafe{} play between the most and least
risk-accepting human groups (approximately five percentage points) lying
within the corresponding uncertainty interval. This null result
independently reproduces \citeauthor{falling_behind_unsafe}'s own
preregistered finding that elicited risk preference did not significantly
predict \Unsafe{} choice \citep{falling_behind_unsafe}, and stands in
contrast to the tournament literature's general finding that human
risk-taking responds to competitive standing rather than remaining a fixed
trait \citep{risk-taking_touraments,risk_taking_in_competition,%
competition_and_risk-taking}: in this task, the round-by-round interaction
captured in \S\ref{sec:results-diversity} and \S\ref{sec:results-position}
is a stronger organising signal than static pre-game disposition. A
weaker, second-order association does survive at the level of behavioural
phenotype rather than mean rate: the same elicited measure differs across
the raw-sequence archetypes of \S\ref{sec:results-diversity} (Kruskal--
Wallis $H=21.95$, $p=0.001$, $n=286$ clustered human players), so the
elicitation may help stratify \emph{which} dynamic pattern a participant
falls into even though it does not predict \emph{how much} \Unsafe{} they
play overall.

\begin{figure*}[t]
    \centering
    \includegraphics[width=\textwidth]{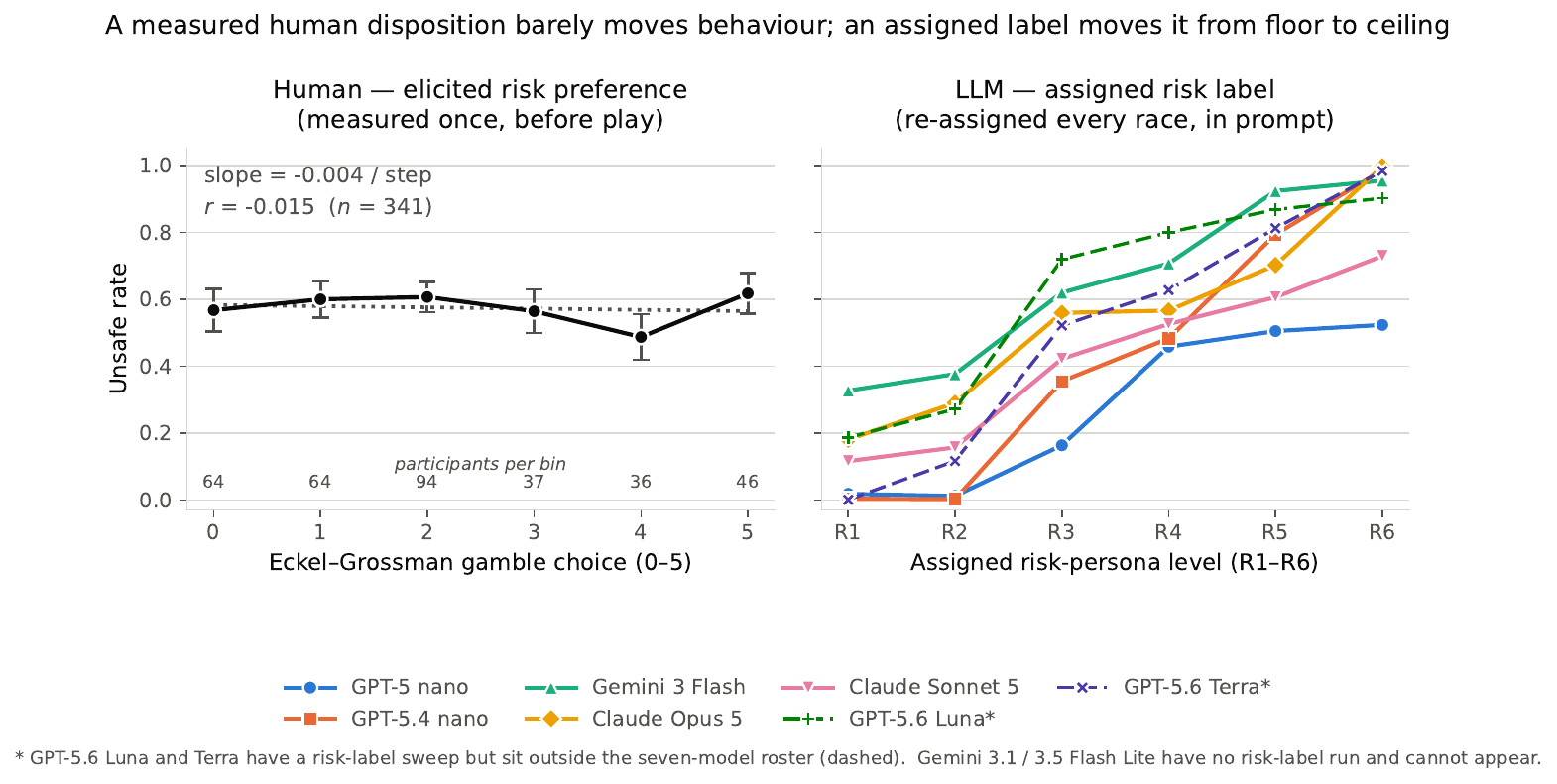}
    \caption{A measured human disposition against an assigned model label.
    Left: mean \Unsafe{} rate by elicited Eckel--Grossman gamble choice,
    with 95\% confidence intervals and participants per bin; the dotted
    line is the trend across the six group means. Right: mean \Unsafe{}
    rate by assigned risk-persona level, for every model with a six-level
    sweep; GPT-5.6 Luna and Terra (dashed) sit outside the seven-model
    roster. The shared six-step axis is visual alignment only --- the left
    is a disposition measured once before play, the right an instruction
    re-assigned every race.}
    \label{fig:own-risk-dependence}
    \Description{Two panels sharing a vertical Unsafe-rate axis and a
    six-step horizontal axis. The left panel plots human mean Unsafe rate
    against the elicited Eckel-Grossman gamble choice from 0 to 5, with 95
    percent confidence intervals and the number of participants beneath
    each point; the six points stay between 0.49 and 0.62 and a dotted
    trend line through them is almost flat. The right panel plots mean
    Unsafe rate against the assigned risk-persona level R1 to R6 for seven
    models, each with its own colour and marker shape. Every model line
    climbs steeply from left to right, from near the floor or lower middle
    at R1 to between 0.52 and 1.0 at R6; the two GPT-5.6 lines are dashed
    because they sit outside the seven-model roster.}
\end{figure*}

Figure~\ref{fig:own-risk-dependence} (left) shows how little that null
leaves. Human mean \Unsafe{} play stays inside a 49--62\% band across the
entire elicitation range, never approaching the low-\Unsafe{} models of
\S\ref{sec:results-diversity} (GPT-5-nano 17\%, Claude Sonnet 5 22\%,
Claude Opus 5 33\%) or the high-\Unsafe{} Gemini models (73--83\%).
Elicited risk preference does not move a participant far enough to resemble
a different model's policy.

The right panel is the opposite (Table~\ref{tab:persona-effect}): every
model with a six-level sweep climbs 50.5 to 98.3 points from R1 to R6, all
but the two nano models monotonically, and those two dip by well under a
point at R2 before rising. The label does not, however, move models by a
common factor --- GPT-5.6 Terra runs lowest at R1 and finishes within 1.3
points of the top, while GPT-5-nano starts fifth of seven and ends last
--- so curve order is not preserved, and the same instruction is worth
very different amounts to different models. Of the seven tested models, five have such a sweep;
neither Gemini Flash-Lite checkpoint does. In logistic models that
additionally control for the mechanism's actual private-risk treatment, the
coefficients on the agent's own assigned risk label remain large for the
three models where that controlled fit has been run, so the persona effect
is not simply standing in for the mechanical risk treatment.

\begin{table}[t]
    \caption{LLM risk-persona effect: change in \Unsafe{} rate from the
    lowest (R1) to the highest (R6) assigned risk label, and the per-level
    OLS slope, for every model with a six-level risk-label sweep ($n=360$
    players per level, $n=330$ for Gemini-3-Flash). The persona coefficient
    (logistic model additionally controlling for the mechanism's actual
    private-risk treatment) has been fit only for the first three; ``--''
    marks models for which only the descriptive delta and slope are
    available. Starred rows sit outside the seven-model roster.}
    \label{tab:persona-effect}
    \centering
    \small
    \begin{tabular}{@{}lrrr@{}}
        \toprule
        Model & $\Delta$ \Unsafe{} (pp) & Per-level slope & Coefficient \\
        \midrule
        GPT-5-nano       & 50.5 & 0.123 & 0.93 \\
        GPT-5.4-nano     & 98.2 & 0.212 & 1.47 \\
        Gemini-3-Flash   & 62.8 & 0.139 & 2.03 \\
        Claude Opus 5    & 81.6 & 0.152 & -- \\
        Claude Sonnet 5  & 61.3 & 0.129 & -- \\
        GPT-5.6 Luna$^*$  & 71.5 & 0.156 & -- \\
        GPT-5.6 Terra$^*$ & 98.3 & 0.203 & -- \\
        \bottomrule
    \end{tabular}
\end{table}

This contrast does not show that LLM agents possess a latent risk
preference comparable to the human elicitation: the human variable is a
measured pre-game disposition that turns out to be nearly unrelated to
in-game behaviour, whereas the LLM variable is an explicit instruction
embedded in the prompt. The more defensible reading is that an assigned
risk persona functions as a high-salience behavioural instruction for
these models, not as an analogue of a stable human risk
disposition. Because the LLM estimates come from a separate risk-matrix
experiment, available for five of the seven tested models and with the
private-risk-controlled logistic fit run for only three of those five,
this comparison remains exploratory and cross-experimental rather than a
like-for-like measurement.

\subsection{Predictive structure of \Unsafe{} choices}
\label{app:predictive-structure}

The preceding section shows that populations produce different action
patterns. We now ask which recorded parts of the game best predict their next
Unsafe choice. This is a study of association, not cause or internal
reasoning. For each population, we fit a random forest, a model that combines
many decision trees. It predicts decisions after round one from five values
known before the choice: the player's previous action, the opponent's previous
action, the progress gap, the assigned maximum private risk, and the round
number. TreeSHAP then measures how much each value contributes to the fitted
model's predictions. A larger SHAP share means that the predictor relied more
on that feature. It does not mean that the feature caused the decision.

\begin{figure}[t]
    \centering
    \includegraphics[width=0.48\textwidth]{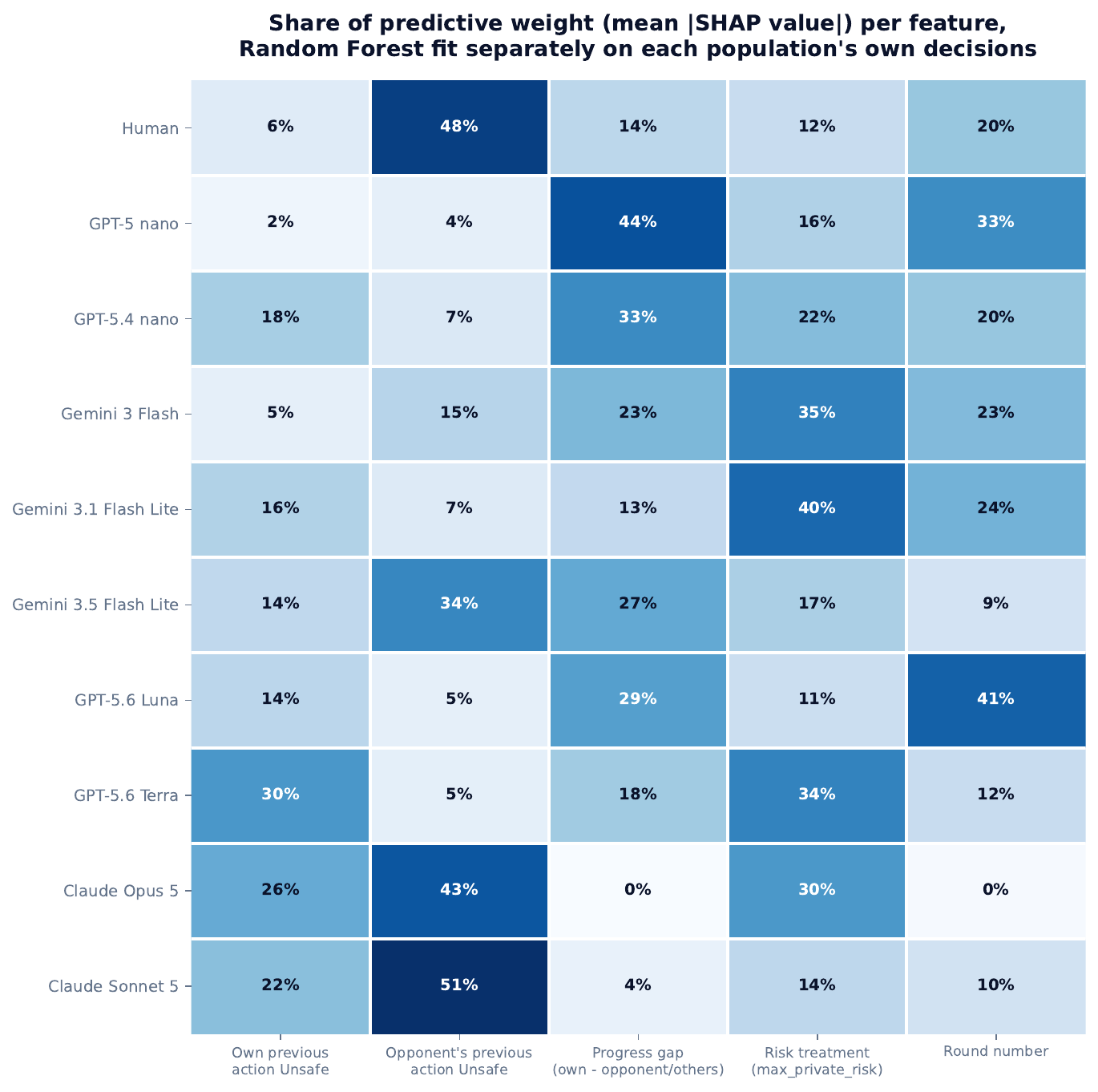}
    \caption{Population-specific predictive structure for round-$t\geq2$
    \Unsafe{} choices. Each row reports the share of mean absolute SHAP
    value assigned to the same five pre-decision features by a random
    forest fitted separately to that population. SHAP magnitudes describe
    how the fitted classifier uses observed features; they are neither
    causal effects nor evidence of the agents' internal reasoning.}
    \label{fig:feature-importance}
    \Description{Heatmap of the share of mean absolute SHAP value that a
    random forest assigns to each of five pre-decision features, with one
    row per population and one column per feature: own previous action,
    opponent's previous action, progress gap, assigned risk treatment, and
    round number. The human row is dominated by the opponent's previous
    action at 48 percent. GPT-5-nano is dominated instead by progress gap
    at 44 percent, the two Gemini Flash checkpoints by the risk treatment
    at 35 and 40 percent, and Claude Sonnet 5 by the opponent's previous
    action at 51 percent.}
\end{figure}

\begin{table*}[t]
    \caption{Population-specific random-forest prediction of \Unsafe{}
    play. Feature columns report shares of total mean absolute SHAP
    magnitude (rows sum to 100\%, up to rounding); AUC and balanced
    accuracy are evaluated on the population-specific test split.}
    \label{tab:feature-importance}
    \centering
    \footnotesize
    \setlength{\tabcolsep}{4.5pt}
    \begin{tabular}{@{}lrrrrrrr@{}}
        \toprule
        Population
        & Own previous
        & Opponent previous
        & Progress gap
        & Risk
        & Round
        & Test AUC
        & Balanced accuracy \\
        \midrule
        Human                   & 6\%  & \textbf{48\%} & 14\% & 12\%          & 20\% & 0.62 & 0.54 \\
        GPT-5-nano              & 2\%  & 4\%           & \textbf{44\%} & 16\% & 33\% & 0.82 & 0.50 \\
        GPT-5.4-nano            & 18\% & 7\%           & \textbf{33\%} & 22\% & 20\% & 0.56 & 0.53 \\
        Gemini-3-Flash          & 5\%  & 15\%          & 23\% & \textbf{35\%} & 23\% & 0.95 & 0.81 \\
        Gemini-3.1-Flash-Lite   & 16\% & 7\%           & 13\% & \textbf{40\%} & 24\% & 0.94 & 0.81 \\
        Gemini-3.5-Flash-Lite   & 14\% & \textbf{34\%} & 27\% & 17\%          & 9\%  & 0.80 & 0.70 \\
        Claude Opus 5           & 26\% & 43\%$^\dagger$ & 0\% & 30\%         & 0\%  & 1.00 & 1.00 \\
        Claude Sonnet 5         & 22\% & \textbf{51\%} & 4\%  & 14\%          & 10\% & 0.96 & 0.92 \\
        \bottomrule
    \end{tabular}

    \vspace{2pt}
    {\footnotesize $^\dagger$Collinear with the risk treatment, not a
    reciprocity estimate; see the discussion below.\par}
\end{table*}

For human participants, the opponent's previous action accounts for 48\%
of total SHAP magnitude, far more than any other feature, while the
player's own previous action contributes only 6\%: a reciprocity-dominated
profile that independently recovers the pattern found in the human
logistic analysis. GPT-5-nano is close to the opposite profile: relative
race position (progress gap) accounts for 44\% of its predictive
importance and opponent history for only 4\%, and its apparently strong
test AUC of 0.82 is misleading on its own, since balanced accuracy is
exactly 0.50: the model plays so close to the \Safe{} floor that the
classifier effectively fails to recover the rare \Unsafe{} class. Gemini-3-
Flash and Gemini-3.1-Flash-Lite are primarily risk-treatment-driven (35\%
and 40\% of SHAP magnitude), consistent with their strong, monotone
descriptive risk-response curves. Claude Sonnet 5 is the only model whose
profile matches the human ordering outright, with the opponent's previous
action largest at 51\%, slightly above the human 48\%;
Gemini-3.5-Flash-Lite is the next closest (34\%), though its progress gap
remains more influential than in the human model. GPT-5.4-nano has no
comparably dominant predictor: importance is
spread across progress gap, risk, round, and its own previous action,
with opponent history at only 7\% and the lowest above-chance predictive
fit in the table (AUC 0.56, balanced accuracy 0.53), so its behaviour is
comparatively difficult to reconstruct from these five mechanical state
variables rather than clearly governed by any one signal.

Claude Opus 5 shows why these shares must be read against the design that
produced them. Its 43\% opponent share reads as strong reciprocity and is
not: on the neutral lane it plays \Unsafe{} on essentially every low-risk
decision and essentially none at high risk, so the risk treatment alone
separates its choices, and since both seats behave that way the opponent's
previous action is a proxy for the risk level rather than an independent
signal. The forest then splits importance between two variables encoding
the same event, and the perfect AUC describes a step function, not a better
fit. A feature-importance share is interpretable only where the feature
keeps variation independent of the others, which is what a
near-deterministic policy removes.

Predictive performance varies substantially across populations (AUC 0.56
to 1.00), so SHAP shares should not be read as directly comparable measures
of decision quality; floor and ceiling behaviour mechanically limits
minority-class prediction for several models. The defensible
conclusion is narrower than ``LLMs use different features'': the same
observable state variables organise predictive information differently
across populations, and matching a human aggregate action rate does not
imply matching the human reciprocity-dominated association profile.

\subsection{Two-player checkpoint descriptive baseline (full scope)}
\label{app:twoplayer-baseline}

The five-checkpoint two-player descriptive baseline referenced throughout
\S\ref{sec:results-diversity} and \S\ref{sec:results-shap} completed 150
races and 2{,}790 decisions, crossing five checkpoints (GPT-5-nano,
GPT-5.4-nano, Gemini-3-Flash-preview, Gemini-3.1-Flash-Lite,
Gemini-3.5-Flash-Lite) with the mechanism's three assigned private-risk
levels ($0.10,0.60,0.90$; Eq.~(2)). At the level of
descriptive curves rather than the pooled aggregate rates already reported
in the main text, GPT-5-nano's \Unsafe{} rate was low and nearly flat
across the three risk levels; GPT-5.4-nano's response to risk level was
non-monotone; and the three Gemini checkpoints' \Unsafe{} rate declined as
the assigned maximum risk increased -- the opposite direction from the two
OpenAI checkpoints. Because provider routes, protocol signatures, and pilot
sample sizes differ across checkpoints, these five curves demonstrate
qualitative heterogeneity in how each checkpoint responds to the risk
manipulation; we do not pool them into a single model-family estimate, and
the formal test of that heterogeneity is the checkpoint-by-risk interaction
reported in \S\ref{sec:results-diversity} ($\chi^2(10)=354.7$).
Table~\ref{tab:twoplayer-baseline-full} gives the per-checkpoint,
per-risk-level rates underlying the qualitative curve descriptions above.

\begin{table}[h]
    \caption{Player-level \Unsafe{} rate by checkpoint and assigned
    maximum-private-risk level, five-checkpoint two-player descriptive
    baseline ($n=20$ players per checkpoint--risk cell, ten races each).}
    \label{tab:twoplayer-baseline-full}
    \centering
    \small
    \begin{tabular}{@{}lrrr@{}}
        \toprule
        Checkpoint & Risk 0.10 & Risk 0.60 & Risk 0.90 \\
        \midrule
        GPT-5 nano             & 12.3\%  & 15.1\%  & 14.5\% \\
        GPT-5.4 nano            & 57.7\%  & 49.6\%  & 56.7\% \\
        Gemini 3 Flash          & 100.0\% & 72.3\%  & 53.9\% \\
        Gemini 3.1 Flash Lite   & 100.0\% & 80.1\%  & 69.9\% \\
        Gemini 3.5 Flash Lite   & 83.8\%  & 70.8\%  & 62.6\% \\
        \bottomrule
    \end{tabular}
\end{table}

\subsection{$N$-player self-play scope ($N=3,4,5$)}
\label{app:nplayer-scope}

\S\ref{sec:results-position} reports position effects for GPT-5-nano and
GPT-5.4-nano at fixed $N$; this subsection gives the broader group-size
scope referenced there. Matched OpenAI self-play baselines cover
$N\in\{3,4,5\}$ for both checkpoints: 180 pilot races, 720 player
trajectories, and 6{,}120 decisions in total, pooled across the mechanism's
three risk levels. Table~\ref{tab:nplayer-scope} reports the aggregate
\Unsafe{} rate by checkpoint and group size.

\begin{table}[h]
    \caption{Aggregate \Unsafe{} rate by group size, OpenAI self-play,
    pooled across the three assigned private-risk levels. Ten independent
    races per model--risk--$N$ cell.}
    \label{tab:nplayer-scope}
    \centering
    \small
    \begin{tabular}{@{}lrrr@{}}
        \toprule
        Checkpoint & $N=3$ & $N=4$ & $N=5$ \\
        \midrule
        GPT-5-nano    & 16.2\% & 26.8\% & 21.9\% \\
        GPT-5.4-nano  & 83.7\% & 87.0\%  & 84.0\% \\
        \bottomrule
    \end{tabular}
\end{table}

Neither checkpoint's response to group size is monotone in $N$, and the
association is checkpoint-specific: GPT-5-nano's rate roughly doubles from
$N=3$ to $N=4$ before partially receding at $N=5$, while GPT-5.4-nano stays
within a narrow 83--87\% band throughout. We do not read either pattern as
an isolated group-size effect, for two reasons. First, changing $N$ also
changes the stage-payoff rule itself (Eq.~(3)--(4) use the joint count
$k^t$ of Safe-choosing companies rather than a single opponent's move),
the number of opponents, and prompt length, so $N$ is confounded with the
representation of the game as well as with its size. Second, the pilots
use only ten independent races per model--risk--$N$ cell, which is a small
base for the per-cell rates entering Table~\ref{tab:nplayer-scope} and
leaves individual cells vulnerable to the same kind of small-$n$ noise
flagged for specific rank $\times$ risk-band cells in
\S\ref{sec:results-position}. Table~\ref{tab:nplayer-scope-full} disaggregates
Table~\ref{tab:nplayer-scope} by assigned risk level and reports a 95\%
Wilson interval for every cell.

\begin{table}[h]
    \caption{\Unsafe{} rate by checkpoint, group size, and assigned
    maximum-private-risk level, OpenAI $N$-player self-play (95\% Wilson
    intervals; $n$ is the decision count per cell).}
    \label{tab:nplayer-scope-full}
    \centering
    \small
    \begin{tabular}{@{}llrrr@{}}
        \toprule
        $N$ & Checkpoint & Risk & $n$ & \Unsafe{} rate (95\% CI) \\
        \midrule
        3 & GPT-5 nano    & 0.10 & 255 & 17.3\% (13.1--22.4\%) \\
        3 & GPT-5 nano    & 0.60 & 255 & 18.0\% (13.8--23.2\%) \\
        3 & GPT-5 nano    & 0.90 & 255 & 13.3\% (9.7--18.1\%) \\
        3 & GPT-5.4 nano  & 0.10 & 255 & 84.3\% (79.3--88.3\%) \\
        3 & GPT-5.4 nano  & 0.60 & 255 & 84.3\% (79.3--88.3\%) \\
        3 & GPT-5.4 nano  & 0.90 & 255 & 82.4\% (77.2--86.5\%) \\
        4 & GPT-5 nano    & 0.10 & 340 & 21.2\% (17.2--25.8\%) \\
        4 & GPT-5 nano    & 0.60 & 340 & 30.3\% (25.7--35.4\%) \\
        4 & GPT-5 nano    & 0.90 & 340 & 28.8\% (24.3--33.9\%) \\
        4 & GPT-5.4 nano  & 0.10 & 340 & 87.6\% (83.7--90.7\%) \\
        4 & GPT-5.4 nano  & 0.60 & 340 & 88.8\% (85.0--91.7\%) \\
        4 & GPT-5.4 nano  & 0.90 & 340 & 84.4\% (80.2--87.9\%) \\
        5 & GPT-5 nano    & 0.10 & 425 & 20.2\% (16.7--24.3\%) \\
        5 & GPT-5 nano    & 0.60 & 425 & 21.2\% (17.6--25.3\%) \\
        5 & GPT-5 nano    & 0.90 & 425 & 24.2\% (20.4--28.5\%) \\
        5 & GPT-5.4 nano  & 0.10 & 425 & 80.2\% (76.2--83.7\%) \\
        5 & GPT-5.4 nano  & 0.60 & 425 & 87.1\% (83.5--89.9\%) \\
        5 & GPT-5.4 nano  & 0.90 & 425 & 84.7\% (81.0--87.8\%) \\
        \bottomrule
    \end{tabular}
\end{table}

\bibliographystyle{ACM-Reference-Format}
\bibliography{references}


\end{document}